\documentclass[12pt]{report} 

\usepackage[margin=1.35in]{geometry}	
\usepackage{algorithm2e}
\RestyleAlgo{ruled}
\usepackage{rotating}
\usepackage{pdflscape}
\usepackage{amsfonts,amsmath,amssymb,wasysym}	
\usepackage{fancyhdr} 							
\usepackage[explicit]{titlesec}	
\usepackage{indentfirst}						
\usepackage[round]{natbib}						
\usepackage{graphicx}
\usepackage{caption}
\usepackage{hyperref}
\usepackage{setspace}
\usepackage[titles]{tocloft}
\usepackage{pdfpages} 
\usepackage{bibentry}
	
\usepackage{times}
\usepackage{url}
\usepackage{latexsym}

\usepackage{graphicx}
\usepackage{tikz}
\usepackage{pgfplots}
\usepackage{subcaption}
\usepackage{url}
\usepackage{booktabs}
\usepackage{microtype}
\usepackage{float}
\usepackage{placeins}

\newcommand{\fix}[1]{\textcolor{blue}{FIXED: #1}}

\SetKwComment{Comment}{/* }{ */}

\hypersetup{ 
    colorlinks=true,
    linkcolor=blue,
    filecolor=magenta,      
    urlcolor=blue,
    pdftitle={UA Dissertation Template}, 
    pdfpagemode=FullScreen,
    citecolor=blue
}
\titleformat{\chapter}{\centering\bfseries}{}{0em}{\chaptertitlename{} \thechapter}[\vspace*{1.5ex}#1]
\titleformat{name=\chapter,numberless}{\centering\bfseries}{}{0em}{#1}
\titleformat{\section}{\bfseries}{}{0em}{\thesection\hspace{1em}#1}
\titleformat{name=\section,numberless}{\bfseries}{}{0em}{#1}
\titleformat{\subsection}{\itshape}{}{0em}{\thesubsection\hspace{1em}#1}
\titleformat{name=\subsection,numberless}{\itshape}{}{0em}{#1}

\makeatletter
\g@addto@macro\appendix{
  \addtocontents{toc}{
    \protect
    \addtolength\cftchapnumwidth{1em}
  }
}
\makeatother

\renewcommand{\listfigurename}{LIST OF FIGURES}
\renewcommand{\listtablename}{LIST OF TABLES}

\newcommand{\CompleteTitle}{Mitigating Data Scarcity For Neural Language Models} 
\newcommand{\FullName}{Hoang Nguyen Hung Van} 
\newcommand{\DegreeType}{Doctor of Philosophy} 
\newcommand{\DepartmentName}{Department of Computer Science} 
\newcommand{\DegreeYear}{2022} 

\begin{document}	
\nobibliography* 


\thispagestyle{empty} 


\vfill

\begin{center}
\MakeUppercase\CompleteTitle \\ 
\vspace*{1.5em}
by\\
\vspace*{1.5em}
\FullName \\
\vspace*{2em}

\rule{3in}{1pt}\\
\vspace*{-1em}
{\small Copyright \copyright\ \FullName\ \DegreeYear} \\
\vspace*{2em}

A Dissertation Submitted to the Faculty of the \\
\vspace*{1.5em}
\MakeUppercase\DepartmentName \\
\vspace*{1.5em}
In Partial Fulfillment of the Requirements\\
\vspace*{0.5em}
For the Degree of\\
\vspace*{1.5em}
\MakeUppercase\DegreeType \\
\vspace*{1.5em}
In the Graduate College\\
\vspace*{1.5em}
THE UNIVERSITY OF ARIZONA\\
\vspace*{3em}
\DegreeYear
\end{center}

\vfill 


\includepdf[pages={1}]{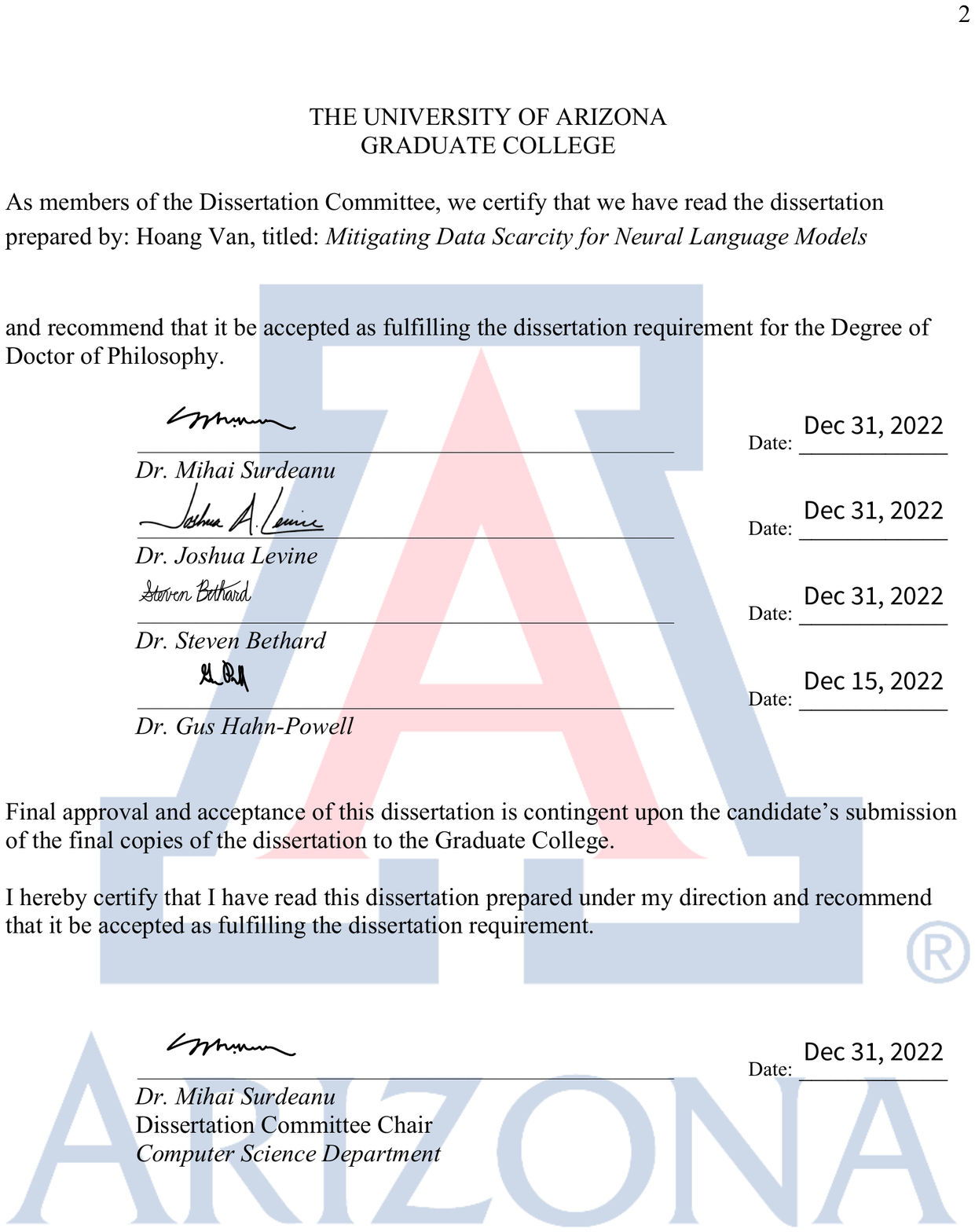} 

\chapter*{STATEMENT BY THE AUTHOR}
\thispagestyle{fancy}

{\flushleft{This dissertation has been submitted in partial fulfillment of requirements for an advanced degree at the University of Arizona and is deposited in the University Library to be made available to borrowers under rules of the Library.}}

{\flushleft{Brief quotations from this dissertation are allowable without special permission, provided that accurate acknowledgment of the source is made. This work is licensed under the Creative Commons Attribution-No Derivative Works 3.0 United States License. To view a copy of this license, visit http://creativecommons.org/licenses/by-nd/3.0/us/ or send a letter to Creative Commons, 171 Second Street, Suite 300, San Francisco, California, 94105, USA.}}

\begin{singlespacing}
\chapter*{ACKNOWLEDGMENTS}
\thispagestyle{fancy} 

First and foremost, my deepest thanks go to my mom to whom I will always bear a debt that I will never be able to repay in full. I want to thank her for tolerating all of my immature stupidities and for being there for me unconditionally.  

I would also like to express my greatest gratitude to my advisor Mihai, who is also a mentor and a role model to me. Since my first interaction with him, I have always been impressed by his utmost care for his students both in terms of their academic success and personal challenges. His research directions and endless lessons have shaped me into the young and enthusiastic researcher as I am now. In the past four and a half years, he has given a treasure of knowledge that I cannot express in writing.

Next, I would like to express my sincere gratitude to the other members of my PhD committee, Steve, Josh, and Gus, for their encouragement, support, and guidance. I learned a lot about how to best visualize data and results so that they will convey impactful and precise meanings. In particular, I would like to thank  Steve for always being willing to provide a second perspective on all of my research work. To Gus, thank you for training me to become a real linguist. Thanks to all professors who have dedicated their time to equipping me with the priceless knowledge that made this dissertation possible and shaped the strong foundation for my research career.

I will always be grateful to Dr. Begoli and his team at Oak Ridge National Laboratories for their valuable guidance and research collaboration opportunities. Their research has given me the motivation of applying NLP advances in the medical field. I will remember all the research discussions and life lessons learned during my time at the lab.

I would like to thank the Thederahn family for being there for me when I needed it the most. Because of their unconditional support, I finally made peace with everything that had happened in my life and found the strength in the very last months of this dissertation.

In addition, I would like to express my gratitude to my marvelous friends, teammates, and collaborators: Zhengzhong, Zheng, Dr. Vikas Yadav, Dr. Mithun Ghosh, Robert, and Masha Thanks for being a part of my PhD journey. I will always remember all your suggestions, research brainstorming sessions, regular happy hours or online meetings filled with chit-chat, gossips, and, laughters. Also, I would like to extend my gratitute to the other labmates in CLULAB at the University of Arizona. Thank you all for making my very first research environment productive, caring, and warming. Every moment we shared was fun, meaningful, and unforgettable.

Also, I want to express my gratefulness for the life-long friendships that I made at Ancestry.com. Thank you all for giving me the opportunity to learn and grow and believing in me unconditionally. I gratefully thank each of them for making positive impacts on my life that I will cherish for a very long time.  

My PhD journey has been incredible, and thank you all for raising me up.

\end{singlespacing}

\chapter*{LAND ACKNOWLEDGMENT}
\thispagestyle{fancy} 

\noindent We respectfully acknowledge the University of Arizona is on the land and territories of Indigenous peoples.
Today, Arizona is home to 22 federally recognized tribes, with Tucson being home to the O'odham and the Yaqui.
Committed to diversity and inclusion, the University strives to build sustainable relationships with sovereign
Native Nations and Indigenous communities through education offerings, partnerships, and community service.

\chapter*{DEDICATION}
\thispagestyle{fancy} 

\begin{center}
\textit{To my grandfather Nguyen Hung Anh, for saving my life that night.}

\textit{Without you, none of these would be possible ...}
\end{center}

\begin{onehalfspacing}
\tableofcontents
\end{onehalfspacing}
\thispagestyle{fancy} 

\cleardoublepage 
\phantomsection 
\addcontentsline{toc}{chapter}{\listfigurename} 
\listoffigures
\thispagestyle{fancy} 

\cleardoublepage 
\phantomsection 
\addcontentsline{toc}{chapter}{\listtablename} 
\listoftables
\thispagestyle{fancy} 

\chapter*{ABSTRACT}
\thispagestyle{fancy} 
\addcontentsline{toc}{chapter}{ABSTRACT}

In recent years, pretrained neural language models (PNLMs) have taken the field of natural language processing by storm, achieving new benchmarks and state-of-the-art performances. These models often rely heavily on annotated data, which may not always be available. Data scarcity are commonly found in specialized domains, such as medical, or in low-resource languages that are underexplored by AI research. In this dissertation, we focus on mitigating data scarcity using data augmentation and neural ensemble learning techniques for neural language models. In both research directions, we implement neural network algorithms and evaluate their impact on assisting neural language models in downstream NLP tasks. Specifically, for data augmentation, we explore two techniques: 1) creating positive training data by moving an answer span around its original context and 2) using text simplification techniques to introduce a variety of writing styles to the original training data. Our results indicate that these simple and effective solutions improve the performance of neural language models considerably in low-resource NLP domains and tasks. For neural ensemble learning, we use a multi-label neural classifier to select the best prediction outcome from a variety of individual pretrained neural language models trained for a low-resource medical text simplification task.

\chapter{Introduction}
\thispagestyle{fancy} 

\section{Motivation}

Natural language processing (NLP) is a branch of artificial intelligence dedicated to devising techniques that enable computers to achieve human-like comprehension of texts/languages. In recent years, pretrained transformer-based \citep{vaswani2017attention} neural language models (PNLMs), such as BERT \citep{devlin2018bert} and T5 \citep{jiang2020neural}, have taken the field of natural language processing by storm, achieving new benchmarks and state-of-the-art performances. These models have strong abilities to extract underlying contextual information encoded in text and languages through expensive pretraining. One way for researchers to utilize pretrained language models' power is to finetune these models specifically for a downstream NLP task of interest. In many cases, these pretrained language models often require large amounts of training data to produce good quality of feature representations\citep{min2021recent}. This thirst for training data might not be always quenched in practice, e.g., in domains and tasks with data scarcity problems. A straightforward solution to data scarcity problems is to annotate more training data. However, this solution is not always feasible due to constraints in human resources and technologies. One good example of these constraints is that, in medical domains, not only is hiring a domain expert for data annotation expensive, but it often requires following strict regulations. Alternatives to this solution are to use data augmentation and ensemble learning techniques to mitigate data scarcity for neural language models \citep{feng2021survey,ganaie2021ensemble}.

In this dissertation, we are particularly interested in how data augmentation and neural ensemble learning can improve the pretrained neural language models' ability to learn downstream NLP tasks well in low-resource domains and tasks. Particularly, our research focus is as follows:

\begin{itemize}
    \item \textbf{Data augmentation:} A set of techniques that increase the amount of training data by generating new data from existing data without manual annotation effort.
    \item \textbf{Neural ensemble learning:} A set of techniques that synthesize training outcomes of more than one neural network trained on the same dataset. 
\end{itemize}

In both research directions, we implement neural network algorithms and evaluate their impact on assisting pretrained neural language models in downstream NLP tasks. Specifically, for data augmentation, we explore two techniques. First, we augment additional positive training data by moving an answer span around its original
context and pairing new answer spans to the original question in low-resource NLP domains and tasks. Second, we propose a novel use of text simplification techniques to introduce a variety of writing styles
to the original training data. For neural ensemble learning, we use a neural classifier to select the best prediction outcome from a variety of PNLMs trained in a low-resource medical domain. 

\section{Contributions}
The key contributions of this dissertation are as follows:

\begin{itemize}
    \item We propose a novel context-based data augmentation to mitigate data scarcity for pretrained neural language models. We show that this simple and effective solution outperforms current state-of-the-art and benchmark performances on TechQA and PolicyQA, two real-world low-resource natural language processing datasets.
    \item In this dissertation, our work further highlights the importance of text simplification control when text simplification techniques are used as a solution for data augmentation. We show that text simplification can assist neural language models in downstream NLP tasks in two different ways: 1) generating additional training data and 2) simplifying input text for machines at inference time. We show that our proposed solutions consistently lead to better performances in a variety of natural language tasks.
    \item Finally, we propose a novel autocomplete approach to medical sentence-level text simplification, which helps reduce the complexity of medical documents while preserving critical medical information.
\end{itemize}

\section{Dissertation Overview}

This dissertation is organized as follows:
\begin{itemize}
    \item In chapter 2, we discuss the background and previous work relevant to our research focus. Specifically, we provide detailed information on transformer architecture and transformer-based pretrained language models. We then discuss previous work on machine reading comprehension as an evaluating NLP task for our data augmentation approach. Background work on text simplification is also provided in this chapter.  
    
    \item In chapter 3, we discuss our first approach to data augmentation: context-based data generation. We then evaluate this approach on two machine reading comprehension datasets, TechQA\citep{castelli2019techqa} and PolicyQA \citep{ahmad2020policyqa}. We present empirical results and discussion of our proposed approach.
    
    \item In chapter 4, we present how to use text simplification methods to augment training data for PNLMs. We start by briefly discussing our motivation and previous work in text simplification. We then provide detailed information about the two datasets, TACRED \citep{zhang2017position} and MNLI \citep{williams2017broad}, used to evaluate our approach. Further, we show that, with a good simplification quality control schema, text simplification techniques can be used to simplify input text at inference time for better model performance. The empirical results of our proposed approach are also presented and discussed in this chapter.
    
    \item In chapter 5, we present our approach of neural ensemble learning to improve PNLM performance on a low-resource medical text simplification task. We first explain how we create a novel parallel dataset for medical text simplification \citep{van2020automets}. We continue by analyzing how PNLMs perform on an autocomplete medical text simplification task. This analysis is then used to create a neural ensemble approach for individual PNLMs for the same task. Finally, We present empirical results and a detailed discussion of our approach. 
    
    \item In chapter 6, we summarize the works that have been presented in this dissertation and discuss future research directions.
\end{itemize}

\chapter{Background \label{chapter:background}} 
\thispagestyle{fancy} 

\section{Transformer-based Pretrained Language Models}

\subsection {Attention Mechanism}
The attention mechanism, which was first introduced in \citep{bahdanau2014neural} is used to capture alignment between an encoder and an decoder in a sequence-to-sequence model. As opposed to a fixed-length encoding vector, the attention mechanism allows the decoder to pay attention to a word/phrase of its choice. Additionally, in this mechanism, each word is a query itself used to generate a list of other words in a sequence that it relates to. This alleviates the need for an encoder alleviates every piece of information available in the source phrase into a tight, fixed-length vector. The attention mechanism was first implemented for the RNN-based encoder and decoder, but it can be reformulated into any sequence-to-sequence architectures as needed. The most successful adaptation of the attention mechanism to a sequence-to-sequence architecture is the transformer architecture, which we will detail next.

\subsection{Transformer Architecture}

Transformer \citep{vaswani2017attention} is an attention-based encoder-decoder architecture. Instead of using general attention mechanism, transformer relies only on a self-attention mechanism which learns representation of a sequence by computing relations of different words in that sequence. This allows transformer to avoid the need for sequence-aligned attention that is usually required in RNN-based architectures. Further, transformer is also implemented with multi-head attention mechanism that comprises multiple single heads of self-attention. In particular, the attention outputs of each single head attention are concatenated to project a final result. The multi-head attention mechanism allows Transformer to extract more well-grounded information from multiple representation spaces, which otherwise would be impossible to extract using a single head attention.

\subsection{Pretrained Language Models}
Utilizing the effectiveness of the transformer attention mechanism and architecture, multiple pretrained natural language networks have been designed to tackle natural language processing tasks. We will detail each of the pretrained language networks used in this dissertation next.

\paragraph{BERT:} Bidirectional Encoder Representations from Transformers  \citep{devlin2018bert} is a method for learning language representations using bidirectional training. BERT was the first of its kind to utilize both masked language modeling (MLM) and next sentence prediction (NSP) task to learn pretrained representations of natural language. In particular, the MLM allows BERT to utilize Transformer architecture to learn bidirectional representation of language. On the other hand, NSP allows BERT to learn the relationship between sentences. This enables BERT to handle tasks that require the understanding of the relationship among sentences, such as question answering and natural language inference. BERT architecture was pre-trained on the BooksCorpus \citep{zhu2015aligning} and English Wikipedia. BERT made state-of-the-arts for 11 NLP benchmarks.

\paragraph{RoBERTa:} A Robustly Optimized BERT Pretraining Approach \citep{liu2019roberta} uses the exact same model architecture as BERT. However, the differences between RoBERTa and BERT are that RoBERTa does not use next sentence prediction loss during pre-training, and further, it uses dynamic masking for masked language modeling. RoBERTa also proposes to pretrain with a larger mini-batch size and larger byte-level byte-pair encoding. Although RoBERTa pretrains on the same text genres, it requires longer sequences and larger training dataset sizes. All in all, RoBERTa provides a more robust optimization for pretraining. The best RoBERTa outperforms BERT across multiple natural language tasks.

\paragraph{XLNet:} Generalized Auto-regressive Pretraining Method \citep{yang2019xlnet}. Like BERT, XLNet benefits from bidirectional contexts. However, XLNet does not suffer the limitations of BERT because of its auto-regressive formulation. This is mostly because XLNet uses permutation language modeling (PLM), where all tokens are predicted in random order. This is the key difference in pretraining of XLNet to that of BERT, where only the masked tokens are predicted. This helps the model learn bidirectional relationships; therefore, better handle dependencies and relations between words. Another difference is that XLNet utilizes Transformer-XL architecture instead of the original Transformer. XLNet made state-of-the-arts for 20 NLP downstream tasks and benchmarks.

\paragraph{GPT-2:} Generative Pretrained Transformer 2 \citep{radford2019language}. Like BERT, GPT-2 is also based on the Transformer network; however, GPT-2 uses an unidirectional left-to-right pre-training process. In particular, GPT-2 was trained on predicting the next word in a given sequence. This allows GPT-2 to learn the latent representation of the language, hence enabling it to better extract useful features for downstream NLP tasks. Similar to BERT, GPT-2 is also trained on raw text with no human-labelled data. GPT-2 is best used for generating texts from a prompt.

\section{Data Augmentation}
\subsection{What is data augmentation?}

Recent years have seen growing research efforts to utilize data augmentation techniques in natural language processing. This is mostly due to the creation of new natural language tasks, the existence of relatively young domains, exploration in low-resource languages, and the increasing popularity of pretrained language models \citep{min2021recent,li2020diverse,xia2019generalized}. However, each of these techniques often requires a large amount of annotated data which might not be feasible due to constraints in resources and technologies \citep{feng2021survey}. In many cases, NLP researchers and scientists come to data augmentation for solutions to mitigate this data scarcity. Data augmentation (DA) is a set of techniques that increase training data without directly annotating more data, aiming to regularize machine learning models from overfitting during training \citep{hernandez2018data,luong2015effective,shorten2019survey}. 

DA techniques are categorized into (1) rule-based and (2) model-based techniques. We will detail each of them next. 

\subsection{Common approaches}

\begin{figure}
    \centering
    \scalebox{0.9}{\includegraphics{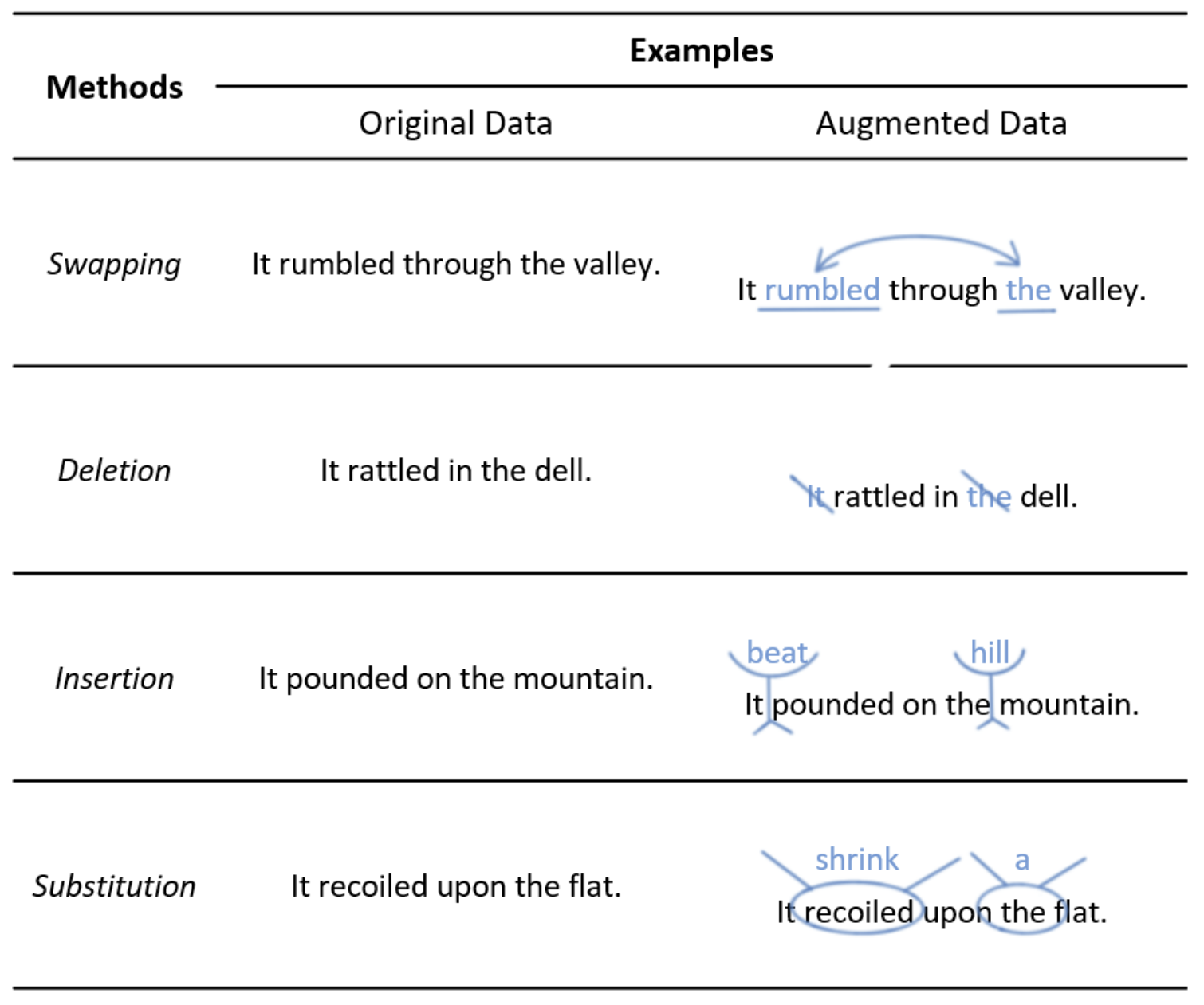}}
    \caption[Visualization of common rule-based techniques for data augmentation in natural language processing.]{Visualization of common rule-based techniques for data augmentation in natural language processing.}
\label{fig:chapter2_rule_based_data_augmentation}
\end{figure}

\paragraph{Rule-based techniques:} This is a set of techniques that relies on hand-crafted transformation rules, including, but not limited to, word replacement, token deletion, and insertion \citep{tang2020revisiting,schwartz2018delta,paschali2019data}. More complex approaches in this family of techniques also include paraphrase identification \citep{chen2020finding} and  dependency-annotated sentences \citep{csahin2019data}. Figure \ref{fig:chapter2_rule_based_data_augmentation} shows common rule-based data augmentation techniques. 

For swapping, some works propose to randomly select two words in a sentence and swap their positions \citep{wei2019eda,longpre2020effective,zhang2020data}. \citep{dai2020analysis} also split text sequence into segments according to labels, and then shuffle the order of tokens inside, with the label order unchanged. In addition to word-level swapping, there are also attempts at sentence-level swapping \citep{luque2019atalaya,yan2019data}. 

For deletion approaches, previous works have randomly removed each word in the sentence with probability p \citep{longpre2019exploration,wei2019eda,zhang2015character,peng2020data}. As for sentence-level deletion, there are attempts at removing random sentences with a certain probability in a legal document. The intuition is that a legal text contains many irrelevant statements, and deleting them will not affect the overall understanding of the legal case \cite{yu2019hierarchical,guo2018dynamic}.

 For insertion, previous works propose to randomly insert synonyms of words that are not stop words in a random position with a certain probability p \citep{shorten2021text,karimi2021aeda,liu2020data}. This can be used to create a noisy variant of the original text input during training. As for sentence-level insertion, in some domains, e.g. legal domains, documents of the same type often share similar sentences. Therefore, adding these sentences will not change the validity of a legal document. The effectiveness of this approach has been first proved in \citep{yan2019data}.

 For substitution, some works propose to replace an original word with another similar word using TF-IDF \citep{xie2020unsupervised,xie2017data,daval2020wmd,lowell2020unsupervised}. In addition to TF-IDF, previous work has created a list of common misspellings in English as a means for word-level substitution and often used the attention mechanism in determining a replacement of the word for data augmentation. \citep{coulombe2018text,regina2020text,xie2017data,qin2020cosda,song2021data,wang2018switchout,daval2020wmd}.

In general, rule-based techniques are easy and quick to compute and augment training data. However, one major drawback is that they are often not agnostic to a task and an underlying model, substantially limiting their ability to generalize.

\paragraph{Model-based techniques:} Sequence-to-sequence and language models have also been used in data augmentation due to their ability to generate text. For example, augmented examples can be generated by replacing words with others randomly drawn according to the recurrent language model’s distribution based on the current context \citep{kobayashi2018contextual}. Previous work also proposes  a task called semantic text exchange, which involves adjusting the overall semantics of a text to fit the context of a new word/phrase that is inserted called the replacement entity \citep{feng2019keep}. Other approaches include syntactic or controlled paraphrasing \citep{iyyer2018adversarial,kumar2020syntax}, document-level paraphrasing \citep{gangal2022nareor}, augmenting misclassified examples \citep{dreossi2018counterexample}, BERT cross-encoder labeling of new inputs \citep{thakur2020augmented}, guided generation using large-scale generative language models \citep{liu2020data}, and automated text augmentation \citep{hu2019learning,cai2020data}. Models can also learn to combine together simpler DA primitives \citep{cubuk2018autoaugment,ratner2017learning} or add human-in-the-loop.

\section{Text Simplification}

The goal of text simplification is to transform text into a variant that is more broadly accessible to a wide variety of readers while preserving the content. Most text simplification techniques fall into two categories: rule-based and sequence-to-sequence. In this section, we will summarize recent advances in the field of text simplification.

\subsection{Simplification is not a summarization.}
One common misperception of text simplification is that text simplification is another form of text summarization, where the goal is to distil important information from the original text. Text simplification is not text summarization. Specifically, text summarization is used to produce a brief summary of the main ideas of the text and often, the summary is significantly shorter than the original text. On the other hand, text simplification aims to reduce the linguistic complexity of the text and retain the original meaning. Therefore, text simplification tends to preserve most to all of the information encoded in the text input. Table \ref{tab:chapter2_text_simplification} shows a comparison of outputs from the two text transformations on the same input text.

\begin{table}
\centering
\scalebox{.96}{%
\begin{tabular}{l l}
\toprule
\textbf{Original text} & Last night, several people were caught to smoke on a flight of China \\
 & United Airlines from Chendu to Beijing. Later the flight temporarily \\
 & landed on Taiyuan Airport. Some passengers asked for a security \\
 & check but were denied by the captain, which led to a collision between \\
 & crew and passengers. \\
\midrule
\textbf{Summarization} & Several people smoked on a flight which led to a collision between \\
 & crew and passengers. \\
\midrule
\textbf{Simplification} & Last night, several people got caught to smoke on a flight of China \\
& United Airlines from Chendu to Beijing. Later the flight landed on \\
& Taiyuan Airport. The captain denied some passengers’ requests for \\
& a security check and a collision between crew and passengers started. \\
\bottomrule
\end{tabular}}
\caption [A comparison of outputs from text simplification and text summarization on the same original text input.]{A comparison of outputs from text simplification and text summarization on the same original text input. As shown, text summarization drops information and keeps only important information encoded in the original text input. On the other hand, the goal of text simplification is to reduce linguistic complexity of the text while preserving original information.}
\label{tab:chapter2_text_simplification}
\end{table}

\subsection{Current approaches for Text Simplification}  

There are, in general, two major types of approaches for text simplification: rule-based  and sequence-to-sequence. The two major approaches are detailed as follows.

\paragraph{Rule-based simplification:} The motivation for rule-based simplification was initially to reduce sentence length as a pre-processing step for a parser \citep{shardlow2014survey}. Recent work has framed text simplification as a two-stage process: 1) analysis followed by 2) transformation. This line of research focuses primarily on syntactic simplification \citep{suter2016rule,evans2019identifying,de2010text}. Specifically, most common approaches are dis-embedding relative clauses and appositives and separating out coordinated clauses. Below is an example of a hand-craft simplification rule \citep{suter2016rule}.

\begin{equation}
    S1 W_{NP}, X_{REL\_PRON} S2, S3. \rightarrow S1 W S3. W S2.
\end{equation}

This rule takes a given text sequence \textit{S1} followed by a noun phrase \textit{W} and a relative pronoun \textit{X} and sequence of words \textit{S2} enclosed in commas, followed by a string \textit{S3}. The rule then splits it into two sentences: \textit{S1 W S3} and \textit{W S2}, with the relative clause \textit{S2} taking \textit{W} as the subject. Table \ref{tab:chapter2_text_simplification_rule_example} shows an example of how this rule can be used in practice. However, these linear pattern-matching rules do not generalize very well. For example, to simplify \textit{``Hoang from Tucson, who was studying at the University of Arizona, played chess, usually on Sundays,''} it is important to decide whether the relative clause attaches to \textit{Hoang} or \textit{Tucson} and whether the clause ends at \textit{Arizona} or \textit{chess}. Hence, disambiguation is challenging for rule-based simplification techniques because they often cannot decide which rules to use when handling test cases like the one discussed above. 

There are also attempts to have a program learn simplification rules from an aligned corpus of sentences. This can be done through aligning parses of a parser on a difficult-simple sentence pair \citep{mandya2014text,mandya2014lexico,de2010text,mandya2014lexico}. During rule learning, these parses are often chunked into phrases. Simplification rules are then induced from a comparison of the structures of the chunked parses of the original and aligned simplification. The objective of the learning algorithm is to flatten sub-trees that appear on both sentences, replacing identical strings of words with variables and then computing tree-to-tree transformations.  

\begin{table}
\centering
\scalebox{1.0}{%
\begin{tabular}{l l}
\toprule
\textbf{Original text} & John, who was the CEO
of a company, played golf. \\
\midrule
\textbf{Simplification} & John played golf. John was the CEO of a company. \\
\bottomrule
\end{tabular}}
\caption [An example usage of a hand-crafted rule for text simplification. Here, an original input text with complex syntactic structure is simplified into two sentences of simpler structures.]{An example usage of a hand-crafted rule for text simplification. Specifically, the original input text with complex syntactic structure is simplified into two sentences of simpler structures.}
\label{tab:chapter2_text_simplification_rule_example}
\end{table}

\paragraph{Sequence-to-sequence simplification:} Recent work has framed test simplification as a monolingual translation task, where the source language needs to be translated into a simplified version of the same language \citep{nisioi2017exploring, nishihara2019controllable,niklaus2017sentence,saggion2017automatic,xu2015problems,xu2016optimizing,jiang2020neural,aluisio2010readability}. This trend has accelerated in recent years due to the availability of the Simple English Wikipedia \footnote{https://simple.wikipedia.org} as a corpus of simplified English. A parallel corpus of aligned difficult (source) and simplified (target) sentences can be created by (a) using Wikipedia revision histories to identify revisions that have simplified the sentence, and (b) aligning sentences in Simple English Wikipedia with sentences from the original English Wikipedia articles. Table \ref{tab:chapter2_text_simplification_dataset} shows an aligned sentence-pair from \citep{coster2011learning}. In chapter \ref{chapter:neural_ensembel_approach}, we introduce a new parallel corpus for medical text simplification, which is in a low-resource domain \citep{van2020automets}. 

\begin{table}
\centering
\scalebox{1.0}{%
\begin{tabular}{l l}
\toprule
\textbf{Original text} & \textbf{Simplified text} \\
\midrule
As Isolde arrives at his side, Tristan dies with & As Isolde arrives at his side, Tristan \\
her name on his lips. &  dies while speaking her name. \\
\midrule
Alfonso Perez Munoz, usually referred to as & Alfonso Perez is a former Spanish \\
Alfonso, is a former Spanish footballer, in the  & football player. \\
striker position. & \\
\midrule
 The reverse process, producing electrical  & A dynamo or an electric generator does \\
 energy from mechanical, energy, is & the reverse: it changes
mechanical \\
 accomplished by a generator or dynamo. &   movement into electric energy. \\
\bottomrule
\end{tabular}}
\caption [Examples of sentence pairs from \citep{coster2011simple}. Each example contains an original text input aligned with the corresponding simplified text input for automated text simplification systems]{Examples of sentence pairs from \citep{coster2011simple}. Each example contains an original text input aligned with the corresponding simplified text input for automated text simplification systems}
\label{tab:chapter2_text_simplification_dataset}
\end{table}

Many recent text simplification systems apply machine translation techniques to learn simplification from these parallel corpora \citep{devaraj2021paragraph,qiang2021lsbert,van2019evaluating}. In general, this line of research is sequence-to-sequence based. They often consist of a translation model, which are a neural language model encoder and decoder. The translation model encodes linguistic information in the original text into and delivers it to the decoder for a final output. The generation is learned through updating training weights of both the encoder and decoder of a translation model. One drawback for this approach is that the generation often relies on the quality of a parallel corpus used for training \citep{xu2015problems}.

There are also hybrid attempts at simplification that combine rules and monolingual translation. Most work in this direction is based on split-and-rephrase transformation and focuses on splitting a sentence into several others, and then making the necessary modifications using a translation model to ensure fluency\citep{reddy2016transforming,narayan2014hybrid,narayan2017split,botha2018learning,rothe2020leveraging}. The systems are controlled so that no deletions are performed to preserve information in the original text input. The authors often use the WEBSPLIT data set \citep{gardent2017creating} to train and test their approaches.

\begin{landscape}
    \begin{figure}
    \centering  \scalebox{0.9}{%
    \includegraphics{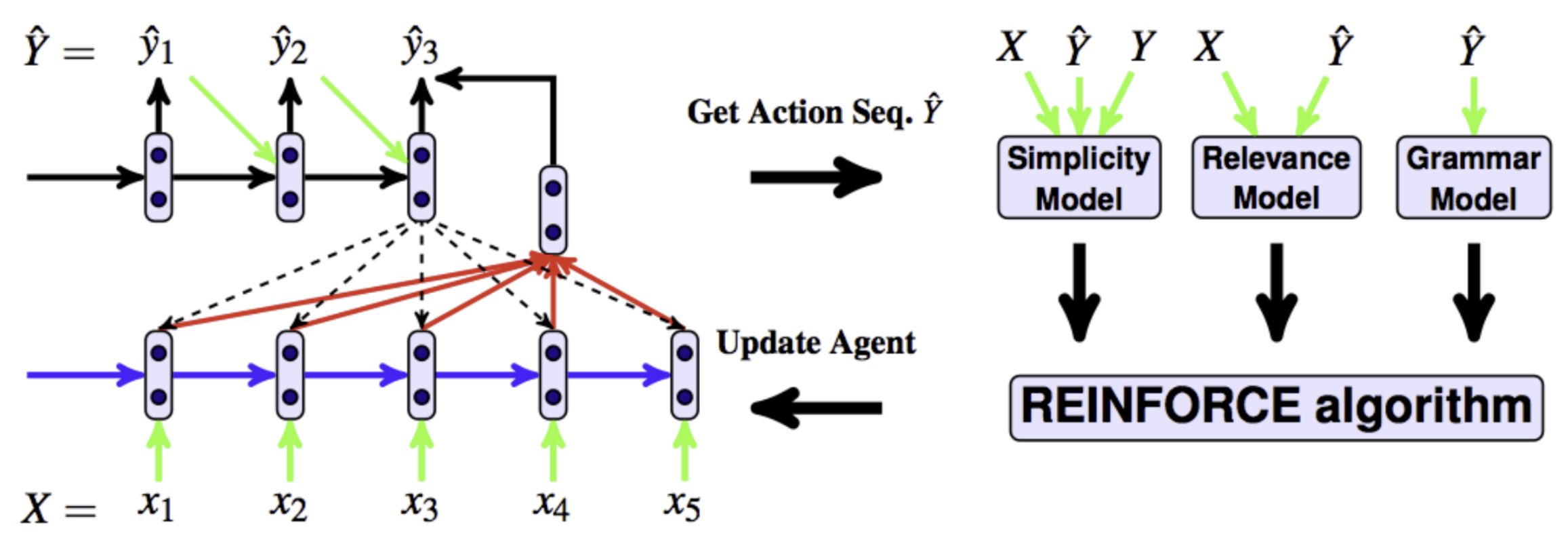}}
    \caption[An overview of the reinforcement learning architecture for text simplification in \citep{zhang2017sentence}. On the left hand side, the approach uses an original transformer-based encoder-decoder architecture as a reinforcement learning agent. The right hand side depicts the training process using text simplification evaluation metrics as a reward.]{An overview of the reinforcement learning architecture for text simplification in \citep{zhang2017sentence}. On the left hand side, the approach uses an original transformer-based encoder-decoder architecture as a reinforcement learning agent. The right hand side depicts the training process using text simplification evaluation metrics as a reward. An advantage of this approach is that the model can be trained end-to-end using metrics specifically designed for simplification.}
    \label{fig:chapter2_text_simplification_rl}
\end{figure}
\end{landscape}

In recent years with breakthroughs of reinforcement learning, interest in utilizing these techniques in the field of text simplification has risen. Specifically, \citep{zhang2017sentence} uses the original transformer-based encoder-decoder architecture as a reinforcement learning agent. Figure \ref{fig:chapter2_text_simplification_rl} shows an overview of this approach. An advantage of this approach is that the model can be trained end-to-end using metrics specifically designed for simplification. In particular, the reinforcement learning agent reads an original text input and takes a series of actions, i.e. words in the vocabulary, to generate the simplified output. After that, it receives a reward that scores the output according to its simplicity, relevance, and fluency. To reward relevance and fluency, some proposed to use cosine similarity scores between the vector representations (obtained using a LSTM or a single neural language model) of the source sentence and the predicted output \citep{nakamachi2020text,lei2021deep,wan2018improving,li2017paraphrase}. However, this reward system introduces new hyperparameters during training. In chapter \ref{chapter:aug_data_ts}, we propose to use the performance on a downstream NLP task as a reward. Specifically, a reinforcement learning agent gets a reward if the generated output makes a correct prediction in a text classification task. For learning, the authors often use the REINFORCE algorithm (Williams 1992), whose goal is to find an agent that maximizes the expected reward. 

\chapter{Simple and Effective Data Augmentation for
Low-Resource Machine Reading \label{chapter:aug_data_qa}} 
\thispagestyle{fancy} 
In recent years, pretrained language models (PNLMs) have taken NLP by storm, achieving state-of-the-art performances on many tasks, including machine reading comprehension. Majority of these PNLMs often rely heavily on annotated data, and therefore, often struggle with data scarcity in low-resource NLP domains and tasks. One straightforward solution to this is to annotate more data. However, it is not always feasible since data annotation is expensive and not an easy task to complete. Particularly, human resources and technology might not be available in every domain. Another solution to mitigating data scarcity is data augmentation which is a more practical approach, as evidenced across NLP tasks and domains \citep{shorten2021text}. In this chapter, we present of our first solution to data scarcity in NLP that is simple and effective. Further, our approach is model-agnostic, making it more applicable to a variety of different neural question answering approaches. Our proposed solution is built upon a data augmentation technique for bounding boxes which is widely used in the field of Computer Vision \citep{dai2015boxsup,zoph2020learning,feng2017dynamic,dvornik2019importance}. Bounding box is a terminology that describes an imaginary rectangular representing a physical object in an image. However, this technique has been rarely used in NLP due to limits coming from the discrete nature of language which makes it more difficult to maintain data invariance. Inspired by this, we propose to first, identify an important text span for a natural language processing task and move this text span around its original location to create additional training data while preserving other information, e.g., label. We will detail this technique in the approach section of this chapter.

In this work, we design a data augmentation technique for text spans that overcomes this limitation of the nature of language for low-resource machine reading comprehension. In particular, by selectively moving a correct answer span, we expand positive training examples for neural question answering systems. The intuition behind our idea is straightforward: rather than directly train a neural approach on the answer spans provided in the training set, we start by pretraining it on the augmented data to identify approximate context where the answer appears. Then we use the pretrained model in two ways. First, instead of training from scratch, we initialize the answer extraction component with weights from this pretrained model before training it. Second, at inference time, we use the pretrained model as an additional document retrieval system. In particular, rather than extracting an answer from any document, we focus only on documents identified as likely to have an answer. Figure \ref{fig:chapter3_intro_example} shows an example of a question in the technical domain from IBM's TechQA dataset\footnote{\url{https://github.com/IBM/techqa}}, for which the current state-of-the-art model predicts incorrectly \citep{castelli2019techqa}, but is correctly predicted by the model trained on our augmented data. Note that, we do not modify the underlying structure of the model, just adding augmented data to the original training dataset. We use random subsampling to select existing data to be augmented. Therefore, our approach is simple and fast. Our approach and empirical results in this chapter are published in the 45th international conference on research and development in information retrieval \citep{van2021cheap}.

The key contributions of our work are three-fold:

\begin{enumerate}
    \item We introduce a simple yet effective data augmentation method for machine reading comprehension. Our approach generates additional training data by artificially moving the boundaries of answer spans in the training partition, and training a model on this data to identify the approximate context in which the answer is likely to appear.
    \item We implement two different strategies to take advantage of the above model. (a) First, we transfer its knowledge into the answer extraction model by initializing with its learned weights, and (b) secondly, at inference time, we use the above model (trained on the fuzzy answer boundaries) to rerank documents based on their likelihood to contain an answer context, and employ answer extraction just in the top documents with the highest scores.
    \item We demonstrate that utilizing approximate reading comprehension knowledge to both document retrieval and answer extraction results in considerable improvements in their performance on low-resource RC tasks. In particular, after transferring the approximate RC knowledge, the BERT-based MRC model's performance is improved by 3.46\%, while the BERT-based document retriever's \citep{castelli2019techqa} performance is improved by 4.33\% (both in absolute F1 score) on TechQA, a real-world low-resource MRC dataset. Further, our best model also outperforms the previous best document retrieval scores \citep{castelli2019techqa} by 15.12\% in document retrieval accuracy (DRA) on TechQA. Similarly, our proposed approach also yields 3.9\% EM and 2.7\% F1 improvements for the BERT-based answer extractor on PolicyQA, another practical and challenging QA dataset that has long answer spans  \citep{ahmad2020policyqa}. Our code base and reproducibility checklist are publicly available\footnote{\url{https://github.com/vanh17/techqa}}.
\end{enumerate}

\section{Chapter Overview}
The remainder of the chapter is organized as follows. We discuss the background work in data augmentation for machine reading comprehension in the next section. In section \ref{sec:chapter3_datasets}, we provide detailed information about the two MRC datasets, TechQA and PolicyQA, and metrics used to evaluate our approach. Our proposed DA approach is explained in section \ref{sec:chapter3_approach}. Section \ref{sec:chapter3_implementation} provides implementation details and the reproducibility checklist. The empirical results of our proposed approach is discussed in section \ref{sec:chapter3_results}. Finally, section \ref{sec:chapter3_conclusion} summarizes the key conclusions of the work presented in this chapter.

\section{Background Work \label{sec:chapter3_background}}

Supervised neural question answering methods have achieved state-of-the-art performance on several datasets and benchmarks \citep{seo2016bidirectional,devlin2018bert,liu2019roberta,yadav2020unsupervised}. However, their success is fueled by large annotated datasets, which are expensive to generate and not always available in low-resource settings. Moreover, machine reading comprehension QA with span selection from a large set of candidate documents has been shown to require even more training data \citep{talmor2019multiqa,fisch2019mrqa,joshi2017triviaqa, dunn2017searchqa}. One straightforward solution to mitigate this need for more training data is to annotate more, however, it might not be often practical in every domain due to constraints in resources and technology \citep{arora2009estimating}. An alternative to that is data augmentation which has been shown to be more feasible across multiple domains \citep{van2001art,shorten2019survey,zoph2020learning}. In recent years, there has been increasing interest in using data augmentation to provide additional data variance for machine reading comprehension systems. There are two main approaches to augment data for machine reading comprehension: rule-based and sequence-to-sequence \citep{shorten2019survey}. Rule-based approaches use transformation rules that often aim for token-level modification. For example, \citeauthor{wei2019eda} proposed to randomly alter tokens in text with a set of operations: deletion, swap, insertion. Similarly, \citep{yun2019cutmix,devries2017improved,ghiasi2020simple} remove part of a text and mix it with a generic patch to create slightly different training data for neural network models.

On the other hand, sequence-to-sequence data augmentation methods modify input text at the sentence level while preserving important information for a downstream task. \textit{BACKTRANSLATION} is a popular choice example for this family of data augmentation methods \citep{sennrich2015neural}. The approach proposed first to translate a text sequence into another language and then back to the original language as a way to create additional data variance. Similarly, PNLMs have been widely used to augment data for machine reading comprehension task \citep{rogelj2018mitigation,yang2020generative,anaby2020not,andreas2019good}. Further, to overcome bias (e.g., class imbalance problem), recent work utilized sequence-to-sequence models to augment misclassified examples \citep{dreossi2018counterexample,radford2019language,thakur2020augmented}.

However, the majority of data augmentation methods that rely on corrupting part or whole text structure are not appropriate for machine reading comprehension with answer span extraction. This is mostly because a corrupted text might often be unrecoverable, and therefore, leading to poor answer span extraction performance by training on these augmented examples \citep{shorten2019survey}. Further, these methods are fairly complicated and computationally expensive \citep{zhong2020random}. Therefore, inspired by data augmentation for bounding boxes in the field of Computer Vision, we design a much simpler and more effective DA method for machine reading comprehension. In particular, we augment data by artificially moving answer spans. Note that, we use random subsampling to select data to be augmented, thus making it cheap to create additional training data for neural network methods.

\section{Datasets and Evaluation Metrics \label{sec:chapter3_datasets}}
\subsection{Datasets}
We had two goals for our analysis: mitigate the lack of training data using data augmentation and domain transfer, and a use of a model trained on fuzzy data to assist in effective dynamic paragraph retrieval. Our models are evaluated on on two complex MRC tasks: TechQA \citep{castelli2019techqa}\footnote{\url{https://github.com/IBM/techqa}} and PolicyQA \citep{ahmad2020policyqa}\footnote{\url{https://github.com/wasiahmad/PolicyQA}}. Both datasets contain questions from {\em low-resource domains}, and their questions are answered by {\em long text spans}, which makes these tasks more challenging for machine reading comprehension approaches \citep{castelli2019techqa}. We will detail each of the two datasets below.

\begin{figure}
\includegraphics[width=1.0\linewidth]
{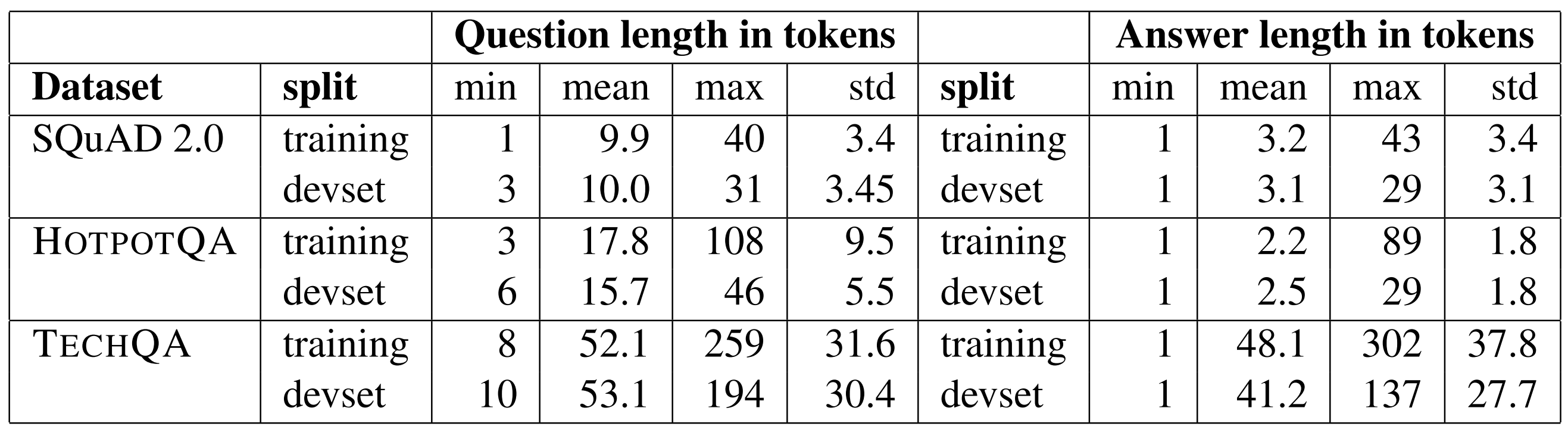}
\caption[Statistics of questions and answer lengths for SQuAD2.0 \citep{rajpurkar2018know}, HotpotQA \citep{yang2018hotpotqa}, and TechQA. Both questions and answers in TechQA are significantly longer in length. Specifically, questions in TechQA are 5.263 and 2.927 times longer than those in SQuAD and HotpotQA, resepectively.]{Statistics of questions and answer lengths for SQuAD2.0 \citep{rajpurkar2018know}, HotpotQA \citep{yang2018hotpotqa}, and TechQA. Both questions and answers in TechQA are significantly longer in length. Specifically, questions in TechQA are 5.263 and 2.927 times longer than those in SQuAD and HotpotQA, resepectively. This proves that TechQA is a more challenging dataset.  and\label{fig:chapter3_length_distribution}}
\end{figure}

\begin{landscape}
    \begin{figure}
    \centering
    \includegraphics[width=1.0\linewidth]{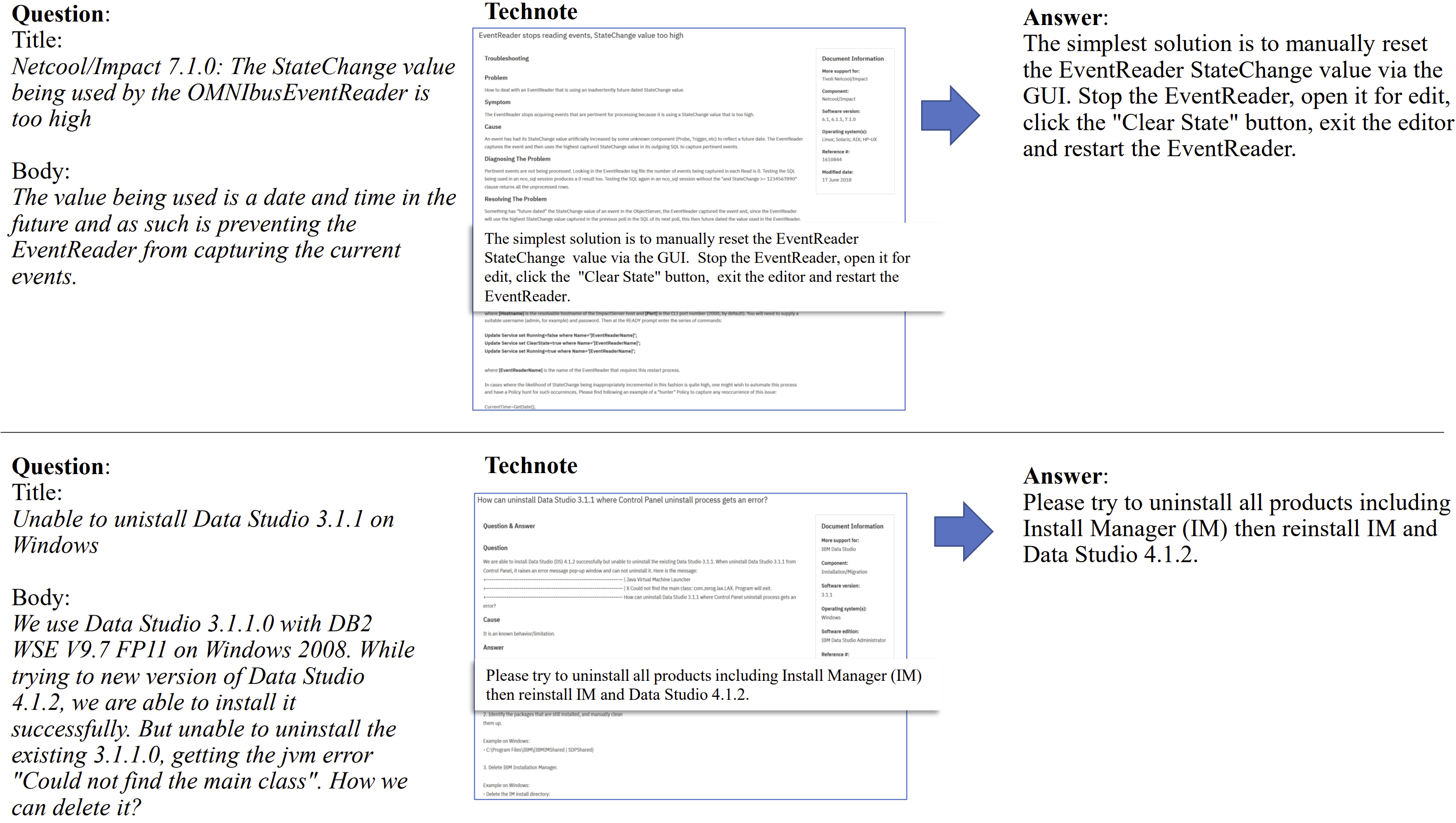}
    \caption[Examples of question-technotes-answer triples in TechQA. The goal of the TechQA task is to retrieve a correct answer span given a technical question and a list of 50 long documents. As shown, each question contains a title and a body, where the title is often a quick summary of the body.]{Examples of question-technotes-answer triples in TechQA. The goal of the TechQA task is to retrieve a correct answer span given a technical question and a list of 50 long documents. As shown, each question contains a title and a body, where the title is often a quick summary of the body. Among 50 documents, there is only \textbf{*one*} document that contains a correct answer span, proving that TechQA is a challenging real-world MRC task.}
    \label{fig:chapter3_example_techqa}
\end{figure}
\end{landscape}

\paragraph{TechQA:} This is a real-world dataset that requires extraction of long answer spans to industrial technical questions. TechQA questions and answers are substantially longer than those found in common datasets \citep{rajpurkar2018know} (see \ref{fig:chapter3_length_distribution}. The average answer span and average question length in the training partition of this dataset are 48.1 and 52.1 tokens, respectively. 
Further, this task is relatively low resourced: the training partition contains only 600 questions; the development one contains 310. Each question is associated with 50 documents that are likely to contain the answer. Those 50 documents are extracted from the ElasticSearch\footnote{\url{https://www.elastic.co/products/elasticsearch}} engine that indexes 801,998 technotes. Note that some questions (25\% in training, 48\% in development) are unanswerable i.e., they do not have an answer in the 50 documents provided. Most questions are categorized in one of the three following groups: 1) generic requests for information, 2) requests for information on how to perform specific operations, and 3) questions about causes and solutions for observed problems. These questions are posted by people who are technically knowledgeable. Further, each question contains a title and a body. In the majority of the questions, a title is a quick summary of the body. Figure \ref{fig:chapter3_example_techqa} shows an example of a question-answer pair with a list of documents. The goal of the task is to retrieve a correct answer span given a question and the provided document list where the answer span is likely to appear.

\begin{figure}
    \centering
    \includegraphics[width=1.0\linewidth]{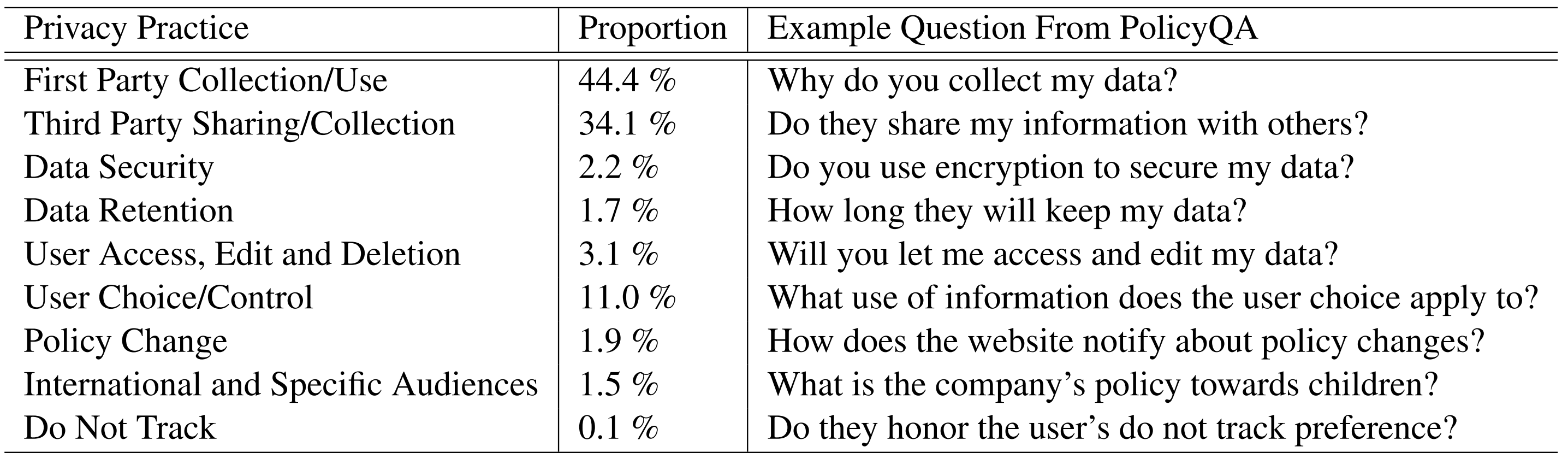}
    \caption[A Question distribution of PolicyQA in privacy practices. There are a total of 9 different practices. The practice with most questions is First Party Collection/use (44\% of all questions). We also provide an example for each privacy policy for reference.]{A Question distribution of PolicyQA in privacy practices. There are a total of 9 different practices. The practice with most questions is First Party Collection/use (44\% of all questions). We also provide an example for each privacy policy for reference. Note that, this figure along with the distribution and examples is provided by the creator of the dataset in the original paper \citep{ahmad2020policyqa}.}
    \label{fig:chapter3_policyqa_categories}
\end{figure}

\paragraph{PolicyQA:} Similar to TechQA, PolicyQA is a another real-world machine reading comprehension task that requires the extraction of answers to privacy policy questions from long documents (average document length is 106 tokens) \citep{ahmad2020policyqa}. However, each PolicyQA question is associated with just 1 given document. PolicyQA is also manually annotated by domain experts. The average text length in PolicyQA is also relatively long: the average answer span and average question length in the training partition of this dataset are 11.2 and 13.3 tokens, respectively. Questions in PolicyQA fall into 9 different privacy practices (see \ref{fig:chapter3_policyqa_categories}).  During annotation, the annotators are provided with the triple \textit{\{Practice, Attribute, Value\}} and the associated text span. For example, given the triple \textit{\{First Party Collection/Use, Personal Information Type, Contact\}} and the associated text span ``name, address,
telephone number, email addres'', the annotators
created questions, such as, (1) What type of contact
information does the company collect?, (2) Will
you use my contact information?, etc. Table \ref{tab:performance_policy} shows an example from PolicyQA. We selected PolicyQA as the second dataset to demonstrate that our approach works with moderate resourced datasets\footnote{PolicyQA has 17K questions in the training partition.} as well.

\subsection{Evaluation metrics} \label{sec:metrics}

\paragraph{Answer extraction:} we directly followed the evaluation metrics proposed by the original task creators \citep{castelli2019techqa, ahmad2020policyqa}. Specifically, we used the following ancillary metrics to evaluate the answer extraction components: F1 (TechQA, PolicyQA) and exact match (EM) (PolicyQA). The F1 score is the macro average over the questions in the evaluation set. 
For each question, the F1 score is computed based on the character overlap between the predicted answer span and the correct answer span.
 We also report the Recall score on TechQA, similarly to the task organizers \citep{castelli2019techqa}.

\paragraph{Document retrieval:} we report the original evaluation metric, i.e., document retrieval accuracy (DRA), as proposed by the task organizers \citep{castelli2019techqa}\footnote{We do not use Precision@$N$ score for retrieval evaluation to keep our results comparable with the other original methods evaluated on these datasets.}.
 In MRC tasks such as TechQA, DRA is calculated only over $answerable$ questions. The score is 1 if the retrieval component retrieves the document containing the correct answer span, and 0 otherwise. We report the performance with DRA@1 and DRA@5, which score the top 1 and 5 retrieved documents, respectively.

\section{Approach \label{sec:chapter3_approach}}

\begin{figure}
\centering
\includegraphics[width=.75\linewidth]
{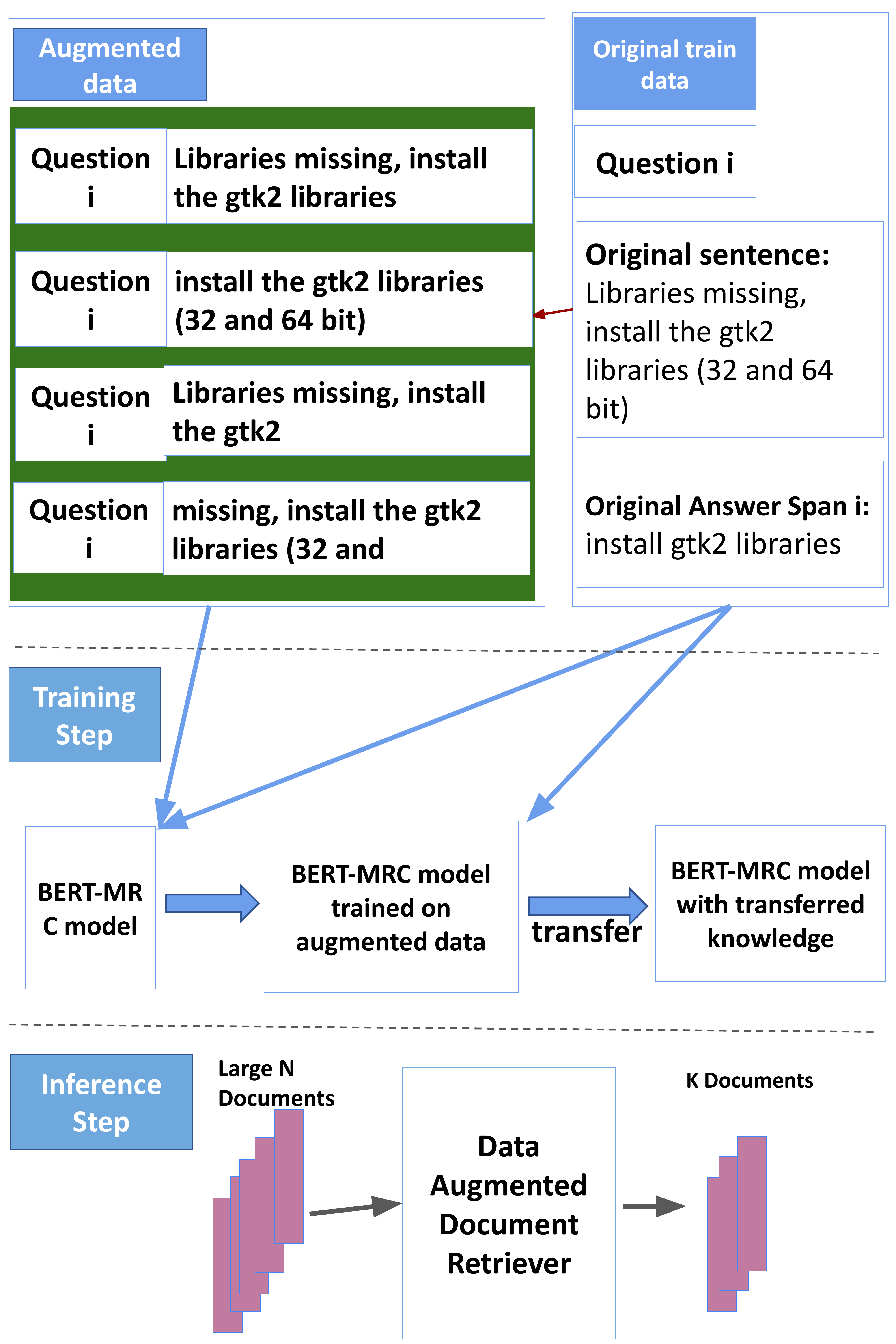}
\caption[Overview of our three key contributions: data augmentation (top), transfer learning (middle), and dynamic document retrieval (bottom). 
Top: we augment question-answer pairs with additional answer spans by moving the answer spans to left or right of the original spans by $d$ characters.]{ Overview of our three key contributions: data augmentation (top), transfer learning (middle), and dynamic document retrieval (bottom). 
Top: we augment question-answer pairs with additional answer spans by moving the answer spans to left or right of the original spans by $d$ characters. 
Middle: we train the MRC answer extraction (AE) component with original settings on the combined augmented + original data. During transfer learning, the models trained on the combined data continue tuning on the original (without augmentation) data.
Bottom: at inference time, $k$ candidate documents are retrieved by the data-augmented document retriever (trained using the scheme above) from the pool of N documents per question. 
\label{fig:model}}

\end{figure}

Fig.\ref{fig:model} highlights the three main components of our machine reading comprehesion approach: (a) data augmentation with fuzzy answer boundaries (top); (b)  transfer learning into the answer extraction component (middle); and (c) improved document retrieval. We detail these three modules next.

\subsection{Data Augmentation} \label{sec:approach_augmentation}
We pose the data augmentation problem \citep{longpre2019exploration, liu2020tell} as a positive example expansion problem. 
First, we sample which answers should be used for augmentation with probability $p$. For example, for $p = 0.3$, 30\% of the answers will be used for augmentation.
Second, once a question-answer pair is selected for augmentation, we generate $n$ additional fuzzy answer spans by moving the correct answer span to the left or right by $d$ characters. This additional answer span is an augmented point and $d$ is an augmentation.
Each generated answer span becomes a new positive training data point. 
For example, for the answer {\em ``install gtk2 libraries''} in the sentence {\em``Libraries missing, install the gtk2 libraries (32 and 64 bit)''}, our data augmentation method generates the spans {\em``Libraries missing, install gtk2 libraries''} ($d = -19$) and {\em``install gtk2 libraries (32 and 64 bit)''} ($d = +16$) as additional answers (see Fig. \ref{fig:model}, data augmentation). 
Both $d$ and the number of augmented data points per training example are hyper parameters that we tune for the given tasks. Note that this augmented data is appended to the original training set.

Intuitively, a model trained on this artificial data (illustrated by the second BERT-MRC block in Figure \ref{fig:model}) would be capable of identifying the \textit{approximate} context in which an answer appears, even if it cannot locate the answer exactly \citep{ golub2017two}.\footnote{This is probably how students work towards answering reading comprehension questions as well. See, e.g.:  \url{https://www.khanacademy.org/test-prep/lsat/lsat-lessons/lsat-reading-comprehension/a/reading-comprehension--article--main-point--quick-guide}.}
given a positive example and a random number ($0 \leq p \leq 1$), the augmentation system (in Fig. \ref{fig:model}) will augment it if the number $p$ is less than a threshold T ($0 \leq T \leq 1$). $T$ can be considered as the number (in percentage) of the positive examples needed to be augmented, i.e., 80\% of the positive examples are augmented if T equals to $0.8$. $T$ is a hyperparameter. The system then, for each example, creates n data points with different window spans in appropriate distances around an original span. We keep other information, such as question title, of the example the same.

\subsection{Transfer Learning} 
As shown in Figure \ref{fig:model}, we first train the BERT-MRC model on the augmented training data consisting of the fuzzy answer spans. Then, we transfer this knowledge of  approximate answer boundaries
by continuing to further train our BERT-MRC model on the original answer spans (without augmentation). The resulting model (the third BERT-MRC block in Fig. \ref{fig:model}) is the final model used for 
 answer span prediction in the complete task. Note that this transfer is trivial: in both situations we use the same neural architecture, BERT-RC \citep{wang2019multi}, and both tasks are formatted in the same way. 
 Note that, while we use BERT-MRC \citep{wang2019multi, wolf2019transformers} to train these answer extraction components, our approach is agnostic to the type of neural architecture used: the 
 same idea could be investigated with other answer extraction methods. 
 

\subsection{Document Retrieval}


Most machine reading comprehension datasets include a list of $N$ candidate documents for each question \citep{kwiatkowski2019natural}. All these documents are scanned for possible answers by typical MRC approaches \citep{swayamdipta2018multi}. Instead, as shown in Fig.~\ref{fig:model},
we rank these documents using the first model trained on the augmented data (with fuzzy answer boundaries), and use only the documents associated with 
top $k$ answers during the answer extraction phase. \fix{For ranking, we use the same document scoring system in \citep{castelli2019techqa}, which assigns maximum answer span score to the corresponding document as a document score.} 
we use only documents that are more likely to contain good answer contexts. In particular, we employ the model trained on the augmented data (with fuzzy answer boundaries) to produce the top $k$ answer spans with the highest scores, and keep only the documents associated with these answer spans. 
For example, on TechQA, by selecting the top documents containing the high score answer spans, our approach narrows down the number of candidate documents to an average of $6.41$ per question\footnote{We also observed that the maximum number of candidate documents is $10$} from the initial pool of 50 documents. As shown in Tab.\ref{tab:performance_policy} 
, feeding a smaller but more relevant pool of candidate documents leads to substantial improvements in the AE performance \citep{yadav2019quick}.

\section{Implementation Details \label{sec:chapter3_implementation}}

For reproducibility, we use the default settings and hyper parameters recommended by the task creators~\citep{castelli2019techqa, ahmad2020policyqa}. Specifically, through this, we aim to separate potential improvements of our approaches from those made with improved configurations. 
Due to computational constraints, we report transfer learning results using BERT Base (uncased) from the HuggingFace library\footnote{\url{https://github.com/huggingface}}.
Additionally, for TechQA, we combine our document retrieval strategy with the answer extraction component based on BERT Large (uncased whole word masking tuned on SQUAD \citep{rajpurkar2018know}) that is provided by the task organizers\footnote{The github repository created by the task organizer can be accessed at \url{https://github.com/IBM/techqa}.}.

\section{Empirical Results and Discussion \label{sec:chapter3_results}}

\begin{table}
\centering
\scalebox{1.12}{%
\begin{tabular}{l l c c c c}
\toprule
& Model & DRA@1 &  DRA@5 & F1 & Recall \\
\midrule
& \textbf{Baselines: Original Settings} & & & &\\
\midrule
1 & BERT Base (OBB) & 31.88 & 45.00 & 51.73 & 51.94 \\
2 & BERT Large (OBL) & 58.13 & 64.38 & 55.31 & 57.81\\
\midrule
& \textbf{Data Augmentation (DA)} & & & & \\
\midrule
3 & 60\% augmentation (aug\_60) & 46.25 & 51.25 & 53.52 & 53.67 \\
4 & 80\% augmentation (aug\_80) & \textbf{47.53*} & \textbf{53.00*} & \textbf{54.35*} & \textbf{54.73*} \\
5 & 100\% augmentation (aug\_100) & 46.33 & 48.75 & 53.56 & 52.76 \\
\midrule
& \textbf{DA+Transfer Learning} & & & & \\ 
\midrule
6 & aug\_60 + Transfer & 43.13 & 50.20 & 54.62 & 55.03\\
7 & aug\_80 + Transfer & \textbf{46.87*} & \textbf{52.10*} & \textbf{55.19*} & \textbf{55.68*} \\
8 & aug\_100 + Transfer & 42.58 & 49.75 & 54.11 & 53.39 \\
\midrule
& \textbf{Document Retrieval + OBL} & & & & \\
\midrule
9 & aug\_60 + OBL & 69.38 & 75.63 & 59.13 & 59.48 \\
10 & aug\_80 + OBL & \textbf{73.25*} & \textbf{78.13*} & \textbf{59.64*} & \textbf{59.84*} \\
11 & aug\_100 + OBL & 71.25 & 73.13 & 58.58 & 58.89 \\
\bottomrule
\end{tabular}}
\caption[Document retrieval accuracies (DRAs) and answer extraction results (F1, Recall) on the development partition of TechQA. We use BERT Base (similar to OBB) for all data augmentation and transfer learning experiments. OBL is a baseline in \citep{castelli2019techqa}.'Data Augmentation' shows results of models tuned with augmented data, starting with pretrained weights from BERT Base.]{\doublespacing Document retrieval accuracies (DRAs) and answer extraction results (F1, Recall) on the development partition of TechQA. We use BERT Base (similar to OBB) for all data augmentation and transfer learning experiments. OBL is a baseline in \citep{castelli2019techqa}.'Data Augmentation' shows results of models tuned with augmented data, starting with pretrained weights from BERT Base. 'Data Augmentation + Transfer Learning' lists results of data augmented models (DAM) after continuing to tune on the original (no augmentation) data. 'Document Retrieval + OBL' uses a hybrid approach where a model from the 'Data Augmentation' block is used to retrieve documents that are then fed to OBL for answer extraction. * indicates significant improvement over the respective baselines (p $<$ 0.02).}
\label{tab:document_retrieveal_accuracy_based_positive_examples}

\end{table}

\begin{table}
\centering
\scalebox{1.}{%
\begin{tabular}{l l c c}
\toprule
& Models & F1 &  EM \\
\midrule
& \textbf{Baselines} & \\
\midrule
1 & OBB pretrained on SQuAD (OBS) & $57.2^{\dag}$ & $27.5^{\dag}$ \\
\midrule
& \textbf{Data Augmentation} & &\\
\midrule
2 & OBS + 3\% augmentation (OBS\_aug\_3) & 58.8 & 29.3 \\
3 & OBS + 4\% augmentation (OBS\_aug\_4) & \textbf{59.8*} & \textbf{30.8*} \\
4 & OBS + 5\% augmentation (OBS\_aug\_5) & 59.2 & 30.2 \\
\midrule
& \textbf{Data Augmentation + Transfer Learning} & & \\
\midrule
5 & OBS\_aug\_3 + Transfer & 59.2 & 29.9 \\
6 & OBS\_aug\_4 + Transfer & \textbf{59.9*} & \textbf{31.4*} \\
7 & OBS\_aug\_5 + Transfer & 59.3 & 30.6 \\
\bottomrule
\end{tabular}}
\caption [Answer extraction results on PolicyQA development. * indicates significant improvement over the baseline (line 1) with p $<$ 0.03. \dag ~indicates experiments from a different paper that we reproduced for a fair comparison. These results differ from the original paper \citep{ahmad2020policyqa}, which is likely due to the different hardware and library versions used.]{ Answer extraction results on PolicyQA development. * indicates significant improvement over the baseline (line 1) with p $<$ 0.03. \dag ~indicates experiments from a different paper that we reproduced for a fair comparison. These results differ from the original paper \citep{ahmad2020policyqa}, which is likely due to the different hardware and library versions used.}
\label{tab:performance_policy}
\end{table}

\begin{sidewaystable}
\scalebox{1.0}{%
\begin{tabular}{l l l l}
\toprule
 & \textbf{Gold Data} & \textbf{Our Approach} & \textbf{Baseline} \\
\midrule
 1 & \textbf{Q:} What is the difference between  & \textbf{Doc:} None, \textbf{Ans:} None & \textbf{Doc:} swg21672885 \\
    & Advanced and AdvancedOnly  & & \textbf{Ans:} The bootstrapProcessServerData ... \\
    & options \textbf{Doc:} None, \textbf{Ans:} & & \\ \\
\midrule
 2 & \textbf{Q:} Is using a monitored JBoss & \textbf{Doc:} swg21967756 & \textbf{Doc:} swg21967756 \\
 
   & a Windows server with ITCAM  & \textbf{Ans:} The JBoss service is & \textbf{Ans:} The JBoss service is not available \\
   
   & supported in  Service? \textbf{Doc:} & not available to run as & \\
   
   & swg21967756 \textbf{Ans:} The JBoss & a Windows service when & \\
   
   & service is not available to & configured with the ITCAM for & \\
   
   & run with the ITCAM ... involves & J2EE agent/DC because this & \\
   
   & changes ... files currently not & involves changes & \\
   
   & supported. & & \\ \\
\midrule
3 & \textbf{Q:} Why do we get server error & \textbf{Doc:} swg21982354 & \textbf{Doc:} swg21981881\\

  & message when running BIRT & \textbf{Ans:} This happens when the   & \textbf{Ans:} In Atlas 6.0.3.3, the Atlas Extensions logging \\
  
  & reports after upgrading to Atlas & BIRT Reports is running in  & configuration has moved to log4j.properties file.\\
  
  & 6.0.3.3? \textbf{Doc:} swg21982354 & Standalone mode and happens & 1. Navigate to Atlas\_Install\_folder/Atlas/Properties \\
  
  & \textbf{Ans:} This happens when the & due to a new configuration - & 2. Edit log4.properties file 3. Update the path  \\
  
  & BIRT Reports is running & report.standalone.userid 1.  & \\
  
  & in Standalone ... & Navigate to Atlas Properties ... & \\
  
  & report.standalone.userid & \#report.standalone.userid=1 & \\
  
\bottomrule
\end{tabular}}
\caption[Qualitative comparison of the outputs from our approach (Data Augmentation + Transfer Learning) and the respective baseline (OBB) on TechQA.] {\doublespacing Qualitative comparison of the outputs from our approach (Data Augmentation + Transfer Learning) and the respective baseline (OBB) on TechQA.}
\label{tab:example_tech}

\end{sidewaystable}

We investigate the contribution of our ideas on document retrieval and AE performance as explained below.

\subsection{Document Retrieval}

Table \ref{tab:document_retrieveal_accuracy_based_positive_examples} shows that data augmentation helps with the document retrieval task in TechQA. Our approaches improve DRA scores considerably compared to the BERT Base baseline (OBB) that serves as the starting point for our experiments (row 1 in Table \ref{tab:document_retrieveal_accuracy_based_positive_examples}).
For example, the 'Data Augmentation' models, which were trained on the data augmented with fuzzy answer boundaries (rows 3--5 in Table \ref{tab:document_retrieveal_accuracy_based_positive_examples}), improve DRA@1 by 14.37--15.65\% (absolute) compared to OBB.
Similarly, the 'Data Augmentation + Transfer Learning' models (rows 6--8 in Table \ref{tab:document_retrieveal_accuracy_based_positive_examples}), which continued to train on the original data, improve DRA@1 by  10.70--14.99\% (absolute).

Interestingly, 'Data Augmentation' models perform better for document retrieval than their equivalent 'Data Augmentation + Transfer Learning' models. For data augmentation and 10.70-14.99\% for transfer learning, both ranges are absolute. Notice that, there are absolute decreases in all DRAs between data augmented models and their transfer learning counterparts (compare pairs of line 3-6, 4-7, and 5-8 in Table \ref{tab:document_retrieveal_accuracy_based_positive_examples}). 
We hypothesize that the transfer learning phase encourages the models to focus on the answer extraction task, ignoring the larger textual context infused in the previous step, which is useful for document retrieval. To further analyze the effects of data augmentation on document retrieval, we also list the results of a hybrid approach, where the documents retrieved by our 'Data Augmentation' models are used as input to the OBL baseline from \citep{castelli2019techqa}. DRA@1 increases noticeably by 11.25--15.12\% (subtract line 2 from lines 9--11 in Table \ref{tab:document_retrieveal_accuracy_based_positive_examples}). We conclude that the candidate documents retrieved by data augmented models can boost the document retrieval performance of an external, independent model.

To illustrate the benefits of additional augmented data, we provide a qualitative comparison between the outputs of our approach and the respective baseline in Table \ref{tab:example_tech}. With the transferred knowledge of larger context, our model successfully retrieves the correct document (row 3 in Table \ref{tab:example_tech}). Further, our method correctly detects if a question has no answer in the given documents (row 1 in Table \ref{tab:example_tech}). The baseline struggles in both scenarios.

\subsection{Answer Extraction}
\paragraph{Data augmented models:} 
As shown in Table \ref{tab:document_retrieveal_accuracy_based_positive_examples}, our data-augmented models (DAMs) yield improved AE performance on TechQA. Compared to the OBB baseline (row 1 in Table \ref{tab:document_retrieveal_accuracy_based_positive_examples}), F1 scores for answer extraction increase when data augmentation is added (see rows 3--5 in Table \ref{tab:document_retrieveal_accuracy_based_positive_examples}).  There are absolute increases of 1.79--2.62\% in F1 for DAMs. Table \ref{tab:example_tech} (row 2) shows a qualitative example where the additional knowledge from the larger context helps the answer extraction component provide better (but not perfect) answer boundaries. Further, there are absolute increases of 3.82\%, 4.33\%, and 3.27\% in F1 scores, respectively, when the DAMs are used to retrieve candidate documents as inputs to OBL for TechQA (rows 9--11 in Table \ref{tab:document_retrieveal_accuracy_based_positive_examples}). 
These results confirm that the better documents retrieved by our method lead to better answers extracted by OBL. Importantly, even though our document retrieval component produces much fewer documents than the original of 50, answer recall increases across the board, further confirming the improved quality of the documents retrieved.

Similarly, on PolicyQA, data augmentation  improves AE performance (see Table \ref{tab:performance_policy}). Compared to the OBS baseline (row 1), evaluation scores for answer extraction increase when data augmentation is added (absolute increases of 1.6--2.6\% and 1.8--3.3\% in F1 and EM, respectively, see rows 2--4)\footnote{Please note that since PolicyQA contains 17,056 questions in its training partition (which is 30  times more than the TechQA training set),  we observed the best improvements on PolicyQA when only 4\% of the training data were used to generate new QA pairs with approximate boundaries.}. 
This suggests that our simple approach is not dataset specific, which is encouraging. Similar to other previous works, we conjecture that our proposed data augmentation approach contributes substantially in very low resource tasks (e.g., TechQA), and its impact decreases as more training resources are available (e.g., PolicyQA). For both datasets, the best performances first increases and then starts decreasing as more data is augmented (see row 3 in Table \ref{tab:performance_policy} and row 4 in Table \ref{tab:document_retrieveal_accuracy_based_positive_examples})\footnote{Note that PolicyQA has around 30  times (17056/600) more training data than TechQA, and therefore, we only need to use the smaller amount of the original data.}. 

\textit{Transfer learning models:} Compared to OBB, our 'Data Augmentation + Transfer Learning' (\texttt{DA+TL}) models perform considerably better for answer extraction in TechQA (rows 6--8 in Table \ref{tab:document_retrieveal_accuracy_based_positive_examples}). Transfer learning models also outperform data augmented models (0.55--1.1\% absolute increases in F1). Comparing the best configurations between the two (rows 4 vs. 7 in Table \ref{tab:document_retrieveal_accuracy_based_positive_examples}), we observe that transfer learning models extract answer spans more effectively (absolute increases of 0.84\% in F1 score). This confirms that transferring to the exact answer spans is beneficial for answer extraction components.

On PolicyQA, our \texttt{DA+TL} models improve AE performance (2.0--2.7\% and 2.4--3.9\% absolute increases in F1 and EM, subtract row 1 from rows 5--7 in Table \ref{tab:performance_policy}). Also, Transfer Learning further improves AE performance (absolute increases of 0.1\% and 0.6\% in F1 and EM, compare the best configurations between the two: rows 4 vs. 7 in Table \ref{tab:performance_policy}). Table \ref{tab:example_policy} shows qualitative examples comparing \texttt{DA+TL} with the OBS baseline. Our qualitative analysis shows that our \texttt{DA+TL} approach more efficiently: (a) locates the correct answers spans (see row 1 in Table \ref{tab:example_policy}), and (b) captures necessary information by expanding the answer span (row 2 in Table \ref{tab:example_policy}).

\section{Conclusion \label{sec:chapter3_conclusion}}

We introduced a simple strategy for data augmentation for machine reading comprehension. Specifically, inspired by data augmentation techniques for boundind boxes in Computer Vision, we generate artificial training data points where answer spans are moved to the left/right of the correct spans. Our apporach used random subsampling to select data to be augmented, making it computationally effective in providing additional data for neural machine reading comprehension methods. Further, we demonstrated that this simple idea is useful for document retrieval (because it was able to capture the larger context in which the answer appears), and also as an effective pretraining step for answer extraction (because of the additional training data, which helps the extraction model narrow its search space). Finally, we showed that our combined contributions yield the best overall performances for two real-world machine reading comprehension datasets, TechQA and PolicyQA, in low-resource domains.

\chapter{Text Simplification as a Solution for Data Augmentation \label{chapter:aug_data_ts}} 
\thispagestyle{fancy} 

In the last chapter, we have described how we utilize the advantages of rule-based data augmentation techniques to mitigate data scarcity problem for pretrained neural language models in machine reading comprehension. Recently, with the broad success of pretrained language models, there has been significant effort in using sequence-to-sequence approaches that transform a text sequence into a variant while preserving some or all encoded information in the original text, to augment data for downstream NLP tasks \citep{nguyen2020improving,hou2018sequence,li2020conditional,weninger2020semi,sugiyama2019data,fadaee2017data,edunov2018understanding,xia2019generalized, corbeil2020bet}. To this end, text simplification, which is a form of text transformation with the goal of reducing text difficulty while preserving important information, has been used as a sequence-to-sequence technique to augment additional data for PNLMs in a variety of NLP tasks \citep{miwa2010entity,schmidek2014improving,wei2019eda,vickrey2008sentence}. Although these various text simplification methods have been shown to generate good augmented data for PNLMs, they are domain-dependent and very expensive to replicate comparable benefits in different natural language tasks and domains \citep{zhang2017sentence,nisioi2017exploring}. In particular, hand-drafted rules are built based on certain syntactic structures that are native to one specific task and not another. We argue that this is mostly due to the rule-based nature of these approaches, where rules are manually and specifically designed for a given natural language task. Also, manual work is needed to design and create these transformation rules, and it is costly to involve human experts in the creation process. More effort dedicated to designing a text simplification approach for data augmentation that is both model- and domain-agnostic is needed. Inspired by this observation, in this chapter, we present a novel application of neural text simplification to data augmentation for PNLMs in a variety of NLP tasks. Our approach and empirical results in this chapter have been accepted for publication at the 2021 conference on empirical methods in natural language processing \citep{van2021may}.

Our contributions are as following:

\begin{enumerate}
    \item We used text simplification to augment training data for two downstream natural language tasks. We empirically analyze this direction using two pretrained neural text simplification systems \citep{martin2019controllable,nisioi2017exploring}, and two tasks: relation extraction using the TACRED dataset \citep{zhang2017position}, and multi-genre natural language inference (MNLI) \citep{williams2017broad} Further, within these two tasks, we explore three neural network methods: two relation extraction approaches, one based on LSTMs \citep{hochreiter1997long} and another based on transformer networks, SpanBERT \citep{joshi2020spanbert}, and one method for MNLI also based on transformer networks, BERT \citep{devlin2018bert}. After augmented data is added, all approaches outperform their respective configurations without augmented data on both TACRED (0.7--1.98\% in F1) and MNLI (0.50--0.65\% in accuracies) tasks.
    \item In the same setting as above, we evaluated text simplification methods in simplifying input texts at inference time for a prediction task. We found that, across two NLP tasks and three neural methods used for these tasks, simplified texts do not often preserve important information needed for downstream NLP models to make correct predictions. We argue that much more effort for controlling the simplification quality is needed for this direction to work. We introduced a simple and effective neural classification approach to control the simplification quality on MNLI using three neural TS systems: ACCESS \citep{martin2019controllable}, MUSS \citep{martin2020muss}, T5 \citep{raffel2020exploring} finetuned on the parallel Wikipedia corpus \citep{jiang2020neural}. Our approach outperforms strong baselines (0.94\% -- 1.78\% in classification accuracies).
\end{enumerate}

\section{Chapter Overview} 
The remainder of the chapter is organized as follows. We discuss the background work in using text simplification methods for data augmentation in the next section. In section \ref{sec:chapter4_datasets}, we provide detailed information about the two datasets, TACRED and MNLI, used to evaluate our approach. Our proposed approach is explained in section \ref{sec:chapter4_approach}. The empirical results of our proposed approach is presented and discussed in section \ref{sec:chapter4_approach}. Finally, section \ref{sec:chapter4_conclusion} summarizes the key conclusions of the work presented in this chapter.

\section{Background Work}
Previous work have effectively proven the practical application of neural networks and neural deep learning approaches to solving machine learning problems \citep{ghosh2021ensemble,blalock2020state,yin2017comparative}. With the recent success of pretrained language models across NLP benchmarks, there has been increased interest in using PNLMs in text simplification \citep{van2020automets,nisioi2017exploring,martin2019controllable,martin2020muss,saggion2017automatic,van2021may}. In this chapter, we focus on two possible directions that TS may help in downstream NLP tasks: 1) simplifying input text at prediction times and 2) augmenting training data for machine. This background work section will be organized to show previous research in the two directions as follows.\ 

\paragraph{Data augmentation by text simplification methods:} Previous work show significant benefits of introducing noisy data on the machine performance \citep{van2021cheap,kobayashi2018contextual}. Previous efforts used TS approaches, e.g. lexical substitution, to augment training data for downstream tasks such as text classification \citep{zhang2015character,wei2019eda}. However, these methods focused on replacing words with thesaurus-based synonyms and did not emphasize other important lexical and syntactic simplification. Here, we use two out-of-the-box neural text simplification systems that apply both lexical and syntactic sentence simplification for data augmentation and show that our data augmentation technique consistently leads to better performances. Note that we do not use rule-based TS systems because they have been proven to perform worse than their neural counterparts \citep{zhang2017sentence,nisioi2017exploring}. Further, rule-based TS systems are harder to build in a domain-independent way due to the many linguistic/syntactic variations across domains.

\paragraph{Input simplification:} For this direction, several works have utilized TS as a pre-processing step for downstream NLP tasks such as information extraction \citep{miwa2010entity,schmidek2014improving,niklaus2017sentence}, parsing \citep{chandrasekar-etal-1996-motivations}, semantic role labelling \citep{vickrey2008sentence}, and machine translation \citep{vstajner2016can}. However, most of them focus on the use of rule-based TS methods. In contrast, we investigate the potential use of domain-agnostic neural TS systems in simplifying inputs for downstream tasks. 
We show that despite the complexity of the tasks investigated and the domain agnosticity of the text simplification approaches, TS improves both tasks when used for training data augmentation, but not when used to simplify evaluation texts.

\section{Datasets \label{sec:chapter4_datasets}}

We evaluate the impact of text simplification on two natural language tasks: (a) relation extraction (RE) using the TACRED dataset \citep{zhang2017position}, and (b) natural language inference (NLI) on the MNLI dataset \citep{williams2017broad}. 

\begin{table}[h]
    \begin{tabular}{l l l}
    \toprule
    & \textbf{Example}     &  \textbf{Entity Types \& Label} \\
    \midrule
    1 & Carey will succeed \textcolor{blue}{Cathleen P. Black}, who held the  & Types: PERSON/TITLE \\
      & position for 15 years and will take on a new role as & Relation: per:title \\
      & \textcolor{red}{chairwoman} of Hearst Magazines, the company said. & \\ \\
    2 & \textcolor{blue}{Irene Morgan Kirkaldy}, who was born and reared in,   & Types: PERSON/CITY \\
      & \textcolor{red}{Baltimore} lived on Long Island and ran a child-care & Relation: per:city of birth \\
      & center in Queens with her second husband, Stanley & \\
      & Kirkaldy. & \\ \\
    3 & \textcolor{red}{Pandit} worked at the brokerage Morgan Stanley for  & Types: ORGANIZATION/ \\
        & about 11 years until 2005, when he and some Morgan & PERSON \\
        & Stanley colleagues quit and later founded the hedge & Relation: org:founded by \\
        & fund \textcolor{blue}{Old Lane Partners.} & \\
    \bottomrule
    \end{tabular}
    \caption[Sampled examples from TACRED dataset for relation extraction. Each sentence contains a subject-object pair. Subject entities are in blue and object entities are in red. For each sentence, there is a relation between the subject and the object.]{Sampled examples from TACRED dataset for relation extraction. Each sentence contains a subject-object pair. Subject entities are in blue and object entities are in red. For each sentence, there is a relation between the subject and the object.}
    \label{tab:chapter4_example_tacred}
\end{table}

\begin{landscape}
    \begin{figure}
    \centering
    \includegraphics[width=1.0\linewidth]{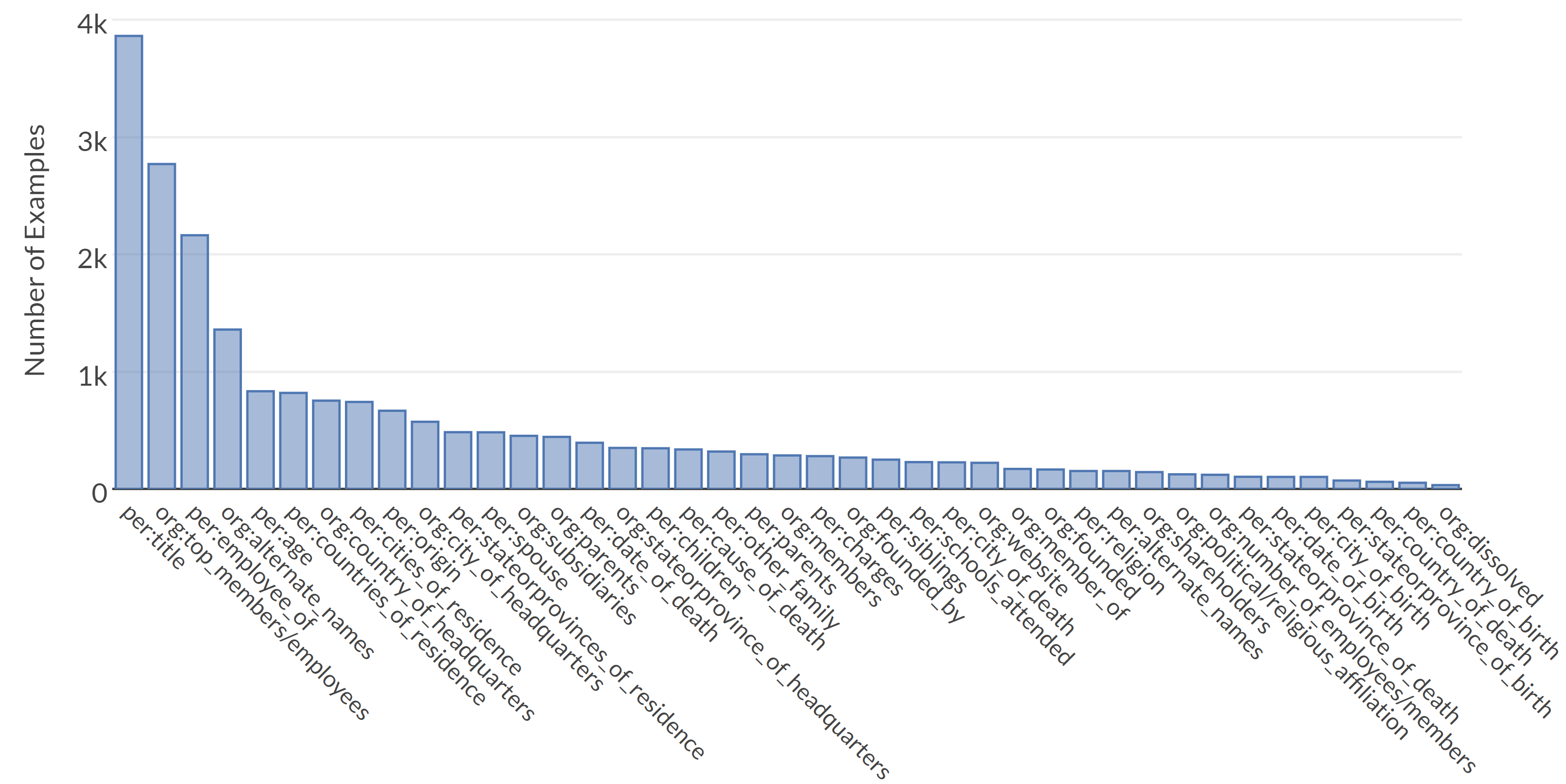}
    \caption[Distribution of relations from examples where a relation was found in TACRED]{Distribution of examples where a relation was found in TACRED. Note that, examples with no relation are manually annotated as \textit{no\_relation} to ensure that models trained on TACRED are not biased towards predicting false positives on real-world text.}
    \label{fig:chapter4_tacred_distribution}
\end{figure}
\end{landscape}

\paragraph{TACRED} is a large-scale relation extraction dataset with 106,264 examples splitted into training(68,124), development(22,631), and test (15,509) sets. The dataset is built on newswire and web text with an average sentence length of 36.4 words. Each sentence contains two entities in focus (called subject and object) and a relation that holds between them. Table \ref{tab:chapter4_example_tacred} shows sampled examples of data from TACRED dataset. Given a sentence with a subject-object pair, the task is to determine a relation between the subject and the object included in the input sentence. To ensure that models trained on TACRED are not biased towards predicting false positives on real-world text, the dataset creators fully annotated all sampled sentences where no relation was found between the mention pairs to be negative examples. As a result, 79.5\% of the examples are labeled as no\_relation. Among the examples where a relation was found, the distribution of relations is shown in figure \ref{fig:chapter4_tacred_distribution}. We selected this task because the nature of RE requires critical information  preservation, which is challenging for neural TS methods \citep{van2020automets}. That is, the simplified sentences must contain the subject and object entities. \newpage

\paragraph{MNLI} is a crowd-sourced collection of 433K sentence pairs annotated for natural language inference (NLI). Natural language inference is the task of determining whether a ``hypothesis'' is true (entailment), false (contradiction), or undetermined (neutral) given a ``premise'' \citep{storks2019recent}. For MNLI, the average sentence length in this dataset is 22.3 words. Each data point contains a premise-hypothesis pair and one of the three labels: contradiction, entailment, and neutral.  The corpus is created from ten different English genres. This helps to approximate the complete diversity of how English is commonly used in the real-world setting. All of the genres appear in the test and development sets, but only five are included in the training set. Therefore, models can be evaluated on both the matched test examples, which are derived from the same sources as those in the training set, and on the mismatched examples, which they have not seen at training time. Table \ref{tab:chapter4_example_mnli} shows examples from MNLI dataset. We selected MNLI as the second task to further understand the effects of text simplification on machine performance on tasks that rely on long text, which is a challenge for TS methods \citep{shardlow2014survey}. 

\begin{table}
    \centering
    \begin{tabular}{l l l}
        \toprule
        \textbf{Premise} & \textbf{Genres \& Label} & \textbf{Hypothesis} \\
        \midrule
        Met my first girlfriend that way. &  FACE-TO-FACE & I did not meet my first \\ 
                                          &  contradiction & girlfriend until later.\\ \\
        Now, as children tend their gardens, they  & LETTERS & All of the children love \\
        have a new appreciation of their & neutral & working in their \\
        relationship to the land, their cultural & & gardens. \\
        heritage, and their community. & & \\ \\
        someone else noticed it and i said well & TELEPHONE & No one noticed and it \\
        i guess that’s true and it was somewhat & contradiction & wasn’t funny at all. \\
        melodious in other words it wasn’t just & & \\
        you know it was really funny & & \\ \\
    \bottomrule
    \end{tabular}
    \caption[Examples from the development set of MNLI dataset. The corpus is derived from ten different English genres, which collectively helps to approximate general ways of how English is commonly used in real-world settings. ]{Examples from the development set of MNLI dataset. The corpus is derived from ten different English genres, which collectively helps to approximate general ways of how English is commonly used in real world.}
    \label{tab:chapter4_example_mnli}
\end{table}

\begin{figure}
    \centering
    \includegraphics[width=0.75\linewidth]{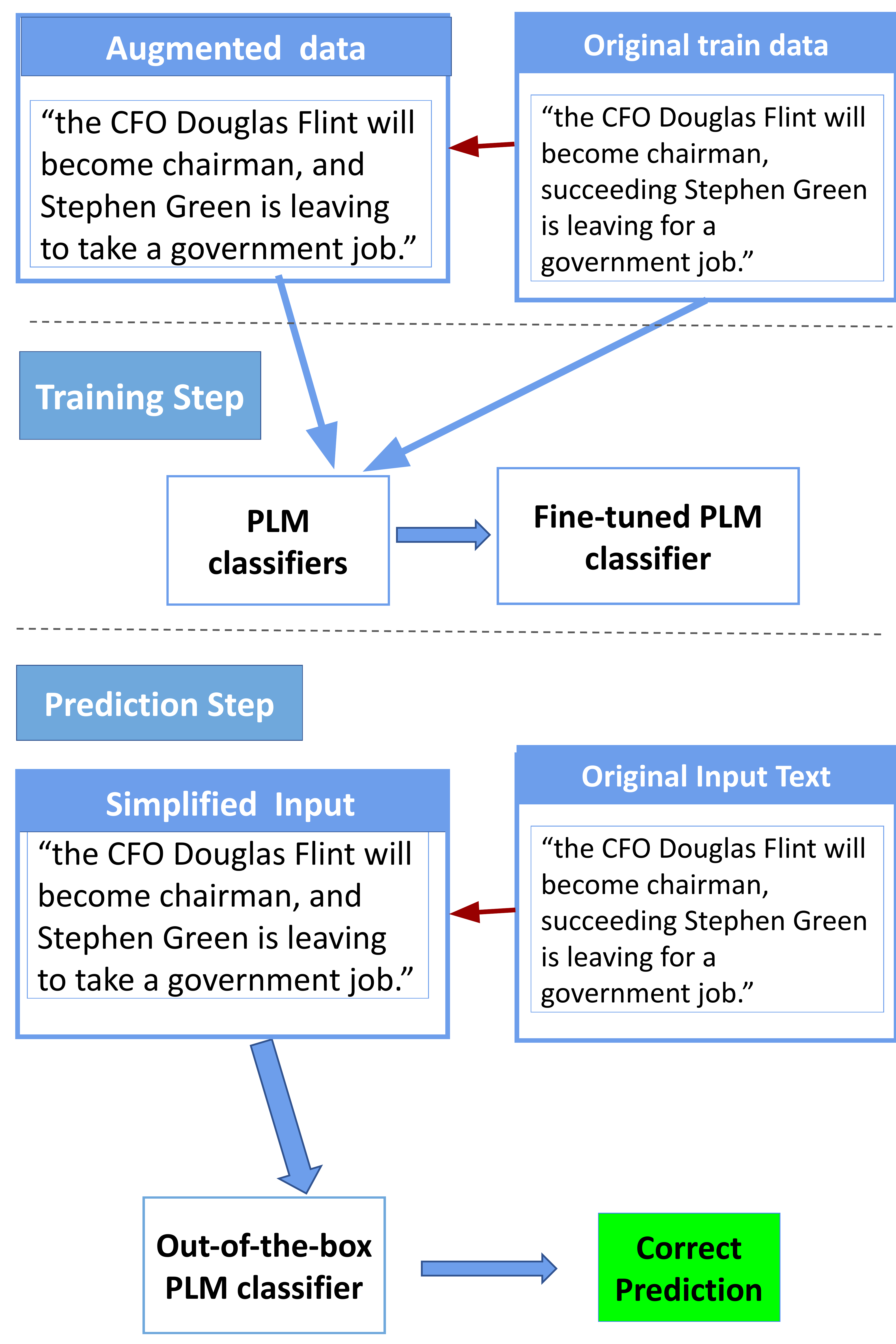}
    \caption[Overview of our key contributions using TS in assisting downstream NLP tasks: data augmentation by text simplification methods (top), and input simplification (bottom)]{Overview of our key contributions using TS in assisting downstream NLP tasks: data augmentation by text simplification methods (top), and input simplification (bottom) Top: we augment additional training data using off-the-shelf text simplification systems. In this setting, we use two TS systems: ACCESS \citep{martin2019controllable} and NTS \citep{saggion2017automatic}.
    Bottom: we simplify input text at inference time for downstream neural network methods. Specifically, instead of making predictions on the original input text, neural models predict based on a simplified text generated by one of the three off-the-shelf TS systems: ACCESS \citep{martin2019controllable}, MUSS \citep{martin2020muss}, T5 \citep{raffel2020exploring} fine-tuned on the Wiki-Auto dataset \citep{jiang2020neural}. Note that, we do not retrain or modify the neural PLM classifier.}
    \label{fig:chapter4_approach}
\end{figure}

\section{Approach \label{sec:chapter4_approach}}

We investigate the impact of text simplification on downstream NLP tasks in two ways: (a) augmenting training data, and (b) simplifying input texts at prediction time for two downstream NLP tasks. Figure \ref{fig:chapter4_approach} visualizes key contributions of our approach. We discuss the settings of these experiments next. 

\subsection{Data Augmentation for Training \label{sec:data_augmentation}}
Here, we augment training data by simplifying the text of some original training examples and appending it to the original training dataset. First, we sample which examples should be used for augmentation with probability $p$. Second, once an example is selected for augmentation, we generate an additional example with the text portion simplified using TS. For example, for the data in section \ref{sec:input_simplification}, we generate additional training data with the corresponding simplified text. $p$ is a hyperparameter that we tuned for each task (see next section).

\begin{table} [h]
\centering
\scalebox{1.0}{%
\begin{tabular}{l l c c c}
\toprule
& & Original & ACCESS & NTS \\
\midrule
1 & Train set & 68,124 & 45,000 (66.06\%) & 49,105 (72.08\%)\\
\midrule
2 & Dev set & 22,631 & 14,451 (63.85\%) & 16,315 (72.09\%) \\
\midrule
3 & Test set & 15,509 & 9830 (63.38\%) & 11,250 (72.54\%) \\
\bottomrule
\end{tabular}}
\caption[Number of data points simplified (by ACCESS \citep{martin2019controllable} and NTS \citep{nisioi2017exploring}) that preserve important information for relation extraction task in TACRED.]{Number of data points simplified (by ACCESS \citep{martin2019controllable} and NTS \citep{nisioi2017exploring}) that preserve important information for relation extraction task in TACRED.}
\label{tab:simplified_tacred_stats}
\end{table}

\paragraph{Text simplification methods:}
For text simplification, we use two out-of-the-box neural seq2seq approaches: ACCESS \citep{martin2019controllable},  and NTS \citep{nisioi2017exploring}. Tables~\ref{tab:tacred_bleu} and \ref{tab:mnli_bleu} show the BLEU scores \citep{papineni2002bleu} between original and simplified text generated by these two TS systems for the two tasks. The tables highlight that both systems change the input texts, with ACCESS being more aggressive. The simplified text is then incorporated to the original datasets with the same format so that we can evaluate the effects of TS on machine performance.

\begin{table}
\centering
\scalebox{1.0}{%
\begin{tabular}{l l c c}
\toprule
& & ACCESS & NTS \\
\midrule
1 & Training Data  &     0.67 $\pm$ 0.16 & 0.89 $\pm$ 0.22  \\
\midrule
2 & Dev Data    & 0.68 $\pm$ 0.15  & 0.92 $\pm$ 0.18  \\
\bottomrule
\end{tabular}}
\caption[The empirical differences in BLEU scores \citep{papineni2002bleu} between original and simplified text generated by two TS systems, ACCESS and NTS, in TACRED training and dev datasets.]{The empirical differences in BLEU scores \citep{papineni2002bleu} between original and simplified text generated by two TS systems, ACCESS and NTS, in TACRED training and dev datasets.} 
\label{tab:tacred_bleu}
\end{table}

\subsection{Input Simplification at Prediction Time \label{sec:input_simplification}}
We pose the run-time input simplification problem as a transparent data pre-processing problem (see bottom of the figure \ref{fig:chapter4_approach}).
That is, given an input data point, we simplify the text while keeping the native format of the task, and then feed the modified input to the actual NLP task.
For example, for the TACRED sentence \emph{``the CFO Douglas Flint will become chairman, succeeding Stephen Green is leaving for a government job.''}, which contains a per:title relation between the two entities {\em Douglas Flint} and {\em chairman},  our approach will first simplify the text to \emph{``the CFO Douglas Flint will become chairman, and Stephen Green is leaving to take a government job.''}. Then we generate a relation prediction for the simplified text using existing relation extraction classifiers.
\begin{landscape}
    \begin{figure}
    \centering    \includegraphics[width=0.9\linewidth]{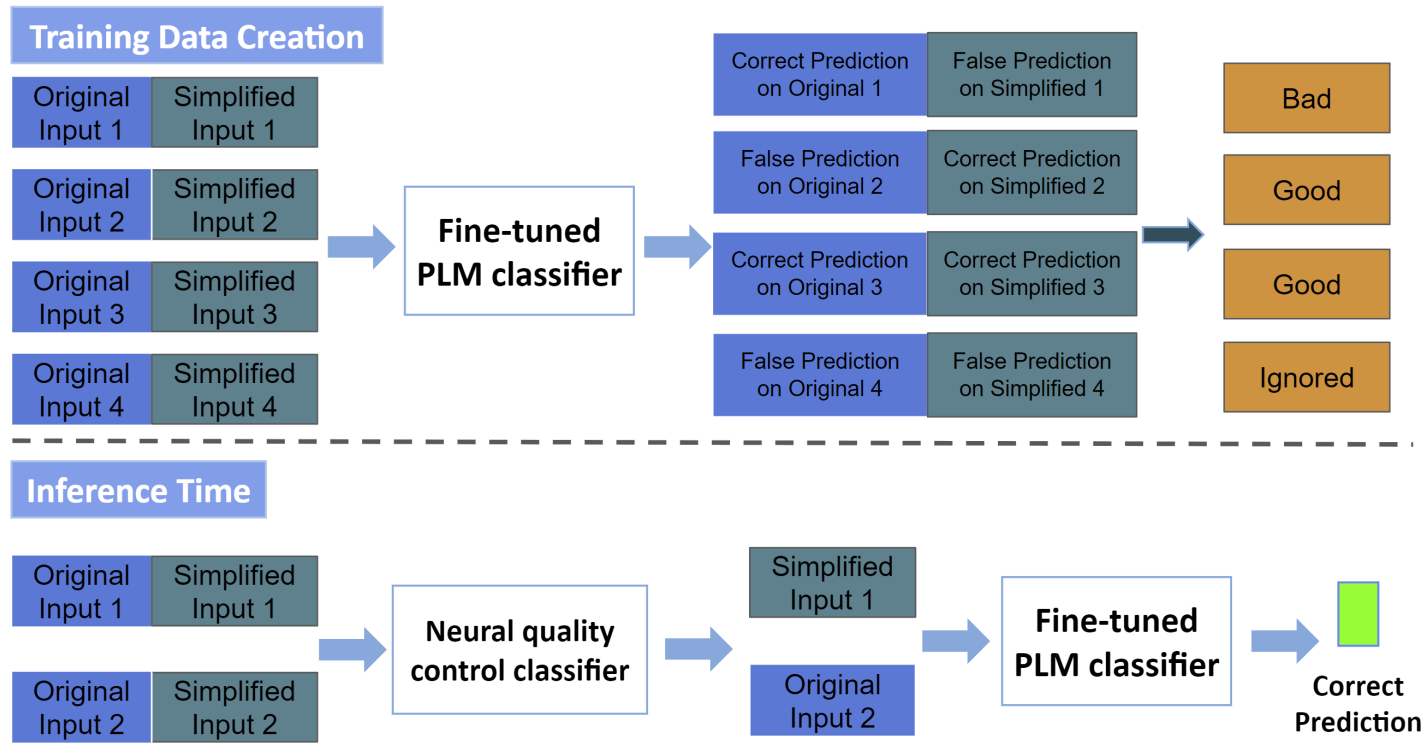}
    \caption[Overview of our text simplification quality control method. Top: the creation of training data for a neural text classifier. Bottom: at inference time, a neural classifier trained on this data is used to determine if a simplified version of an input is selected for a prediction task.]{Overview of our text simplification quality control method. Top: the creation of training data for a neural text classifier. We will use this classifier to determine which text input, i.e., original vs simplified, to use at inference time.  Bottom: at inference time, a neural classifier trained on this data is used to determine if a simplified version of an input is selected for a prediction task. Note that, we do not modify the underlying structure of the fine-tuned PNLM classifier that is used to address a downstream task.}
    \label{fig:chapter4_qc_training_data}
\end{figure}
\end{landscape}

\paragraph{Text simplification quality control method at inference time:} We introduced a novel technique to control simplification quality using feedback from a downstream natural language task. Note that, this technique is used at inference time as opposed to the other simplification quality control method for training data augmentation described before. In particular, we framed a text simplification quality control problem as a text classification problem, i.e., for a given original input text, we will train a neural classifier to identify if the corresponding simplified version is a good replacement at inference time. Figure \ref{fig:chapter4_qc_training_data} shows the training data creation for this classification problem. In particular, we assign a label for an original-simplified pair as follows: a ``good'' label is assigned to a sentence pair if the simplified text helps the underlying model make a correct prediction, and ``bad'' for ones that do not. We ignored a pair where the underlying model makes a wrong prediction on both the original and simplified text. At inference time, we use a simplified text as input to the prediction task only if a neural simplification quality control classifier assigns a ``good`` label to it, otherwise, we select the original.

\section{Empirical Results and Discussion \label{sec:chapter4_results}}

\begin{table}
\centering
\scalebox{1.0}{%
\begin{tabular}{l l c c}
\toprule
& & ACCESS & NTS \\
\midrule
& \bf{\emph{Train}}  &         &       \\
\midrule
1 & Premise    & 0.62 $\pm$ 0.24  &  0.76 $\pm$ 0.25 \\ 
\midrule
2 & Hypothesis & 0.62 $\pm$ 0.30 & 0.80 $\pm$ 0.17 \\
\midrule
& \bf{\emph{Dev mismatched}} &  &  \\
\midrule
3 & Premise    &    0.62 $\pm$ 0.28 & 0.80 $\pm$ 0.22 \\
\midrule
4 & Hypothesis & 0.65 $\pm$ 0.23 & 0.81 $\pm$ 0.17 \\
\midrule
& \bf{\emph{Dev matched}} &  &  \\
\midrule
5 & Premise    &  0.62 $\pm$ 0.30 & 0.75 $\pm$ 0.26 \\
\midrule
6 & Hypothesis & 0.60 $\pm$ 0.25 & 0.80 $\pm$ 0.17 \\
\bottomrule
\end{tabular}}
\caption[The empirical differences in BLEU scores \citep{papineni2002bleu} between original and simplified text generated by two TS systems, ACCESS and NTS, in MNLI training and dev datasets.]{The empirical differences in BLEU scores \citep{papineni2002bleu} between original and simplified text generated by two TS systems, ACCESS and NTS, in MNLI training and dev datasets.} 
\label{tab:mnli_bleu}
\end{table}

\begin{table}
\centering
\scalebox{1.0}{%
\begin{tabular}{l l c c c}
\toprule
& & Simplified & Original & Simplified + \\
& &          &           & Complement \\
\midrule
& \bf{\emph{Train Data Sets}}  &         &       \\
\midrule
& \bf{LSTM}    &         &       \\  
\midrule
1 & Original & 59.95 & 62.70 & 61.32\\
\midrule
2 & Simplified & 57.93 & 56.94 & 57.47\\
\midrule
3 & Simplified +  & 62.34 & 62.59 & 62.12\\
 &  Complement & & & \\
\midrule
4 & Simplified +  & \bf{62.64} & \bf{64.52} & \bf{64.08} \\
  &  Original (AD)          &       &       & \\
\midrule
& \bf{SpanBERT}    &         &       \\  
\midrule
5 & Original & 62.42 & 66.70 & 64.12 \\
\midrule
6 & Simplified & 64.72 & 64.14 & 64.29 \\
\midrule
7 & Simplified +  & 64.12 & 65.45 & 64.92 \\
 &  Complement & & & \\
\midrule
8 & Simplified +  & \bf{65.14} & \bf{68.00} & \bf{66.43} \\
  &  Original (AD)          &       &       & \\
\bottomrule
\end{tabular}}
\caption[F1 on TACRED test set of the LSTM and SpanBERT approaches using ACCESS \citep{martin2019controllable} as the TS method. The different rows indicate the different data augmentation strategies applied on the training data, while the columns indicate the type of simplification applied at runtime on the test data]{\doublespacing F1 on TACRED test set of the LSTM and SpanBERT approaches using ACCESS \citep{martin2019controllable} as the TS method. The different rows indicate the different data augmentation strategies applied on the training data, while the columns indicate the type of simplification applied at runtime on the test data. We investigated the following configurations:
{\em Original}: unmodified dataset; 
{\em Simplified}: dataset completely simplified using TS; 
{\em Simplified + Complement}: consists of simplified data that preserves critical information combined with original data when simplification fails to preserve important information; {\em Simplified + Original}: consists of all original data augmented with additional simplified data that preserves critical information. {\em (AD)} annotates models using data augmented by neural TS systems during training.} 
\label{tab:results_tacred_access}
\end{table}

\begin{table}
\centering
\scalebox{1.0}{%
\begin{tabular}{l l c c c}
\toprule
& & Simplified & Original & Simplified + \\
& &          &           & Complement \\
\midrule
& \bf{\emph{Train Data Sets}}  &         &       \\
\midrule
& \bf{LSTM}    &         &       \\  
\midrule
1 & Original & 60.47 & 62.70 & 61.03\\
\midrule
2 & Simplified & 58.23 & 60.68 & 59.52\\
\midrule
3 & Simplified +  & \bf{63.40} & 62.96 & 62.28 \\
 &  Complement & & & \\
\midrule
4 & Simplified +  & 62.91 & \bf{64.68} & \bf{64.35} \\
  &  Original (AD)          &       &       & \\
\midrule
& \bf{SpanBERT}    &         &       \\  
\midrule
5 & Original & 62.20 & 66.70 & 63.90 \\
\midrule
6 & Simplified & 62.39 & 63.00 & 62.09 \\
7 & Simplified +  & 64.12 & 65.32 & 63.92 \\
 &  Complement & & & \\
\midrule
8 & Simplified +  & \bf{65.32} & \bf{67.40} & \bf{65.47} \\
  &  Original (AD)           &       &       & \\
\bottomrule
\end{tabular}}
\caption[F1 on TACRED test set of the LSTM and SpanBERT approaches using NTS \citep{nisioi2017exploring} as the TS method.]{\doublespacing F1 on TACRED test set of the LSTM and SpanBERT approaches using NTS \citep{nisioi2017exploring} as the TS method. The different rows indicate the different data augmentation strategies applied on the training data, while the columns indicate the type of simplification applied at runtime on the test data. We investigated the following configurations: {\em Original}: unmodified dataset; {\em Simplified}: dataset completely simplified using TS; {\em Simplified + Complement}: consists of simplified data that preserves critical information combined with original data when simplification fails to preserve important information; {\em Simplified + Original}: consists of all original data augmented with additional simplified data that preserves critical information. {\em (AD)} annotates models using data augmented by neural TS systems during training.} 
\label{tab:results_tacred_nts}
\end{table}

Tables~\ref{tab:results_tacred_access} and \ref{tab:results_tacred_nts} summarize our results on TACRED for the two distinct TS methods. Because we tuned the hyperparameter $p$ for MNLI, we report results on both development and test for this task (Tables~\ref{tab:results_mnli_dev} and \ref{tab:results_mnli_test}, respectively). Further, for MNLI we also report average performance (and standard deviation) for 3 runs, where we select a different sample to be simplified in each run. This is not necessary for TACRED; for this task, we simplified {\em all} data points that preserved critical information i.e., the two entities participating in the relation.\footnote{This is not possible for MNLI, where it is unclear which part of the text is critical for the task.}

\subsection{Augmenting training data:} As shown in rows 3 and 6 in Table~\ref{tab:results_tacred_access} and \ref{tab:results_tacred_nts}, all methods trained on augmented data yield consistent performance improvements, regardless of the RE classifier used (LSTM or SpanBERT) or TS method used (ACCESS or NTS). 
There are absolute increases of 1.30--1.82\% F1 for ACCESS and 0.70--1.98\% F1 for NTS (subtract row 1 from row 3 and row 4 from row 6 for ACCESS and NTS respectively). 
The best configuration is when the original training data is augmented with all data points that could be simplified while preserving the subject and object of the relation (rows 4 and 8 in the two tables).
These results confirm that TS systems can provide additional, useful training information for RE methods. 

\begin{table}
\centering
\scalebox{.82}{%
\begin{tabular}{l l c c}
\toprule
& & Simplified & Original \\
& & m/mm acc   & m/mm acc   \\
\midrule
& \bf{\emph{Train Data Sets}}  &         &       \\
\midrule
& \bf{ACCESS}    &   &   \\ 
\midrule
1 & Original & 71.25/71.43 & 82.89/83.10 \\
\midrule
2 & Simplified & \bf{73.49/73.81} & 81.41/81.52 \\
\midrule
3 & Original Swapped   & 71.76 $\pm$ 0.13/ & 83.00 $\pm$ 0.03/ \\
 & with 10\% Simplified & 72.12 $\pm$ 0.08 & 83.25 $\pm$ 0.05 \\
\midrule
4 & Original Swapped   & 72.74 $\pm$ 0.10/ & 82.66 $\pm$ 0.07/ \\
 & with 20\% Simplified & 73.10 $\pm$ 0.12 & 82.88 $\pm$ 0.09 \\
\midrule
5 & 5\% Simplified & 71.30 $\pm$ 0.15/ & \bf{83.47} $\pm$ \bf{0.04} \\
  & + Original (AD) & 71.52 $\pm$ 0.10 & \bf{83.61} $\pm$ \bf{0.08}\\
\midrule
6 & 10\% Simplified  & 71.81 $\pm$ 0.07/ & 82.81 $\pm$ 0.05/ \\
  & + Original (AD)      & 71.99 $\pm$ 0.08 & 83.05 $\pm$ 0.09 \\
\midrule
7 & 15\% Simplified  & 71.87 $\pm$ 0.11/ & 82.92 $\pm$ 0.05/ \\
  & + Original (AD)  & 72.10 $\pm$ 0.07 & 83.13 $\pm$ 0.06 \\
\midrule
8 & 20\% Simplified  & 72.21 $\pm$ 0.05/ & 82.75 $\pm$ 0.09/ \\
  & + Original       & 72.39 $\pm$ 0.04  & 83.05 $\pm$ 0.10 \\
\midrule
& \bf{NTS}    &     &  \\  
\midrule
9 & Original & 33.36/33.53 & 82.89/83.10 \\
\midrule
10 & Simplified & \bf{34.85/35.10} & 82.13/82.35 \\
\midrule
11 & Original Swapped   & 33.39 $\pm$ 0.10/ & 83.28 $\pm$ 0.07/ \\
 & with 10\% Simplified & 33.46 $\pm$ 0.08 & 83.50 $\pm$ 0.11 \\
\midrule
12 & Original Swapped   & 33.71 $\pm$ 0.08/ & 82.60 $\pm$ 0.14/ \\
 & with 20\% Simplified & 33.90 $\pm$ 0.11/ & 82.79 $\pm$ 0.09 \\
\midrule
13 & 5\% Simplified & 33.35 $\pm$ 0.10/ & 83.20 $\pm$ 0.09/ \\
  & +  Original (AD) & 33.50 $\pm$ 0.09 & 83.41 $\pm$ 0.10 \\
\midrule
14 & 10\% Simplified & 33.50 $\pm$ 0.07/ & \bf{83.51} $\pm$ \bf{0.05}/ \\
  &  + Original (AD) & 33.80 $\pm$ 0.09 & \bf{83.70} $\pm$ \bf{0.07} \\
\midrule
15 & 15\% Simplified  & 33.65 $\pm$ 0.04/ & 83.09 $\pm$ 0.05 \\
  &  + Original (AD)  & 33.79 $\pm$ 0.10  &  83.25 $\pm$ 0.07\\
\midrule
16 & 20\% Simplified  & 33.68$\pm$0.08/ & 82.68$\pm$0.10/ \\
  &  + Original      & 33.85$\pm$0.09 & 82.88$\pm$0.15 \\
\bottomrule
\end{tabular}}
\caption[Matched (m) and mismatched (mm) accuracies on MNLI development 
using text simplified/augmented by ACCESS (top half) and NTS (bottom half).
{\em Original Swapped with x\% Simplified} consists of original data with x\% of data points replaced with their simplified form.]{Matched (m) and mismatched (mm) accuracies on MNLI development 
using text simplified/augmented by ACCESS (top half) and NTS (bottom half).
{\em Original Swapped with x\% Simplified} consists of original data with x\% of data points replaced with their simplified form. {\em x\% Simplified + Original} consists of the original data augmented with an additional x\% of simplified data. {\em (AD)} annotates models using data augmented by neural TS systems during training. Note that our results in the original configuration differ slightly from those in \citep{devlin2018bert}. This is likely due to the different hardware and library versions used \citep{belz2021systematic}.}
\label{tab:results_mnli_dev}
\end{table}

Similarly, on MNLI, the classifier trained using augmented data outperforms the BERT classifier which is trained only on the original MNLI data. For two TS systems, ACCESS and NTS, we observe performance increases of 0.59--0.65\% matched accuracy, and 0.50--0.62\% mismatched accuracy (compare rows 1 vs. 2, and row 3 vs. 4 in Table \ref{tab:results_mnli_test}). This confirms that TS as data augmentation is also useful for NLI.

All in all, our experiments suggest that our data augmentation approach using TS is fairly general. It does not depend on the actual TS method used, and it improves three different methods from two different NLP tasks. Further, our results indicate that our augmentation approach is more beneficial for tasks with lower resources (e.g., TACRED), but its impact decreases as more training data is available (e.g., MNLI).

\begin{table}
\centering
\scalebox{1.0}{%
\begin{tabular}{l l c c}
\toprule
& & Simplified & Original \\
& & m/mm acc   & m/mm acc   \\
\midrule
& \bf{\emph{Train Data Sets}}  &         &       \\
\midrule
& \bf{ACCESS}    &   &   \\  
\midrule
1 & Original & 71.10/71.30 & 82.78/83.00 \\
\midrule
2 & 5\% Simplified + & \bf{71.21}/\bf{71.40} & \bf{83.37}/\bf{83.50} \\
 & Original (AD) & & \\
\midrule
& \bf{NTS}    &      \\ 
\midrule
3 & Original & 33.25/33.45 & 82.78/83.00 \\
\midrule
4 & 10\% Simplified +  & \bf{33.39}/\bf{33.61} & \bf{83.43}/\bf{83.62} \\
 & Original (AD) & & \\
\bottomrule
\end{tabular}}
\caption[Matched (m) and mismatched (mm) accuracies on MNLI test, using the best configurations from development.]{ Matched (m) and mismatched (mm) accuracies on MNLI test, using the best configurations from development.} 
\label{tab:results_mnli_test}
\end{table}

\subsection{Input simplification at prediction time:} Tables~\ref{tab:results_tacred_access} and \ref{tab:results_tacred_nts} show that simplifying all inputs at test time does not yield improvements (compare the {\em Original} column with the third one).
There are absolute decreases in performance of 1.38--2.58\% and 1.67--2.80\% in F1 on TACRED for ACCESS and NTS systems, respectively (subtract column 3 from column 2 in rows 1 and 4). 

Similarly, on MNLI, the performance when simplifying all inputs is lower than the classifier tested on the original data. The performance drops on MNLI are more severe (11.68--49.53\% and 11.70--49.55\% in matched and mismatched accuracies) (subtract column 1 from column 2 in row 1 and row 3 in Table \ref{tab:results_mnli_dev} pairwise). We hypothesize that this is due to the quality of simplifications in MNLI being lower than those in TACRED. In the latter situation we could apply a form of quality control, i.e., by accepting only the simplifications that preserve the subject and object of the relation. 
To illustrate the benefits/dangers of text simplification, we show a few examples where simplification improves/hurts MNLI output in Table~\ref{tab:example_mnli}. We restrict the use of only simplifications that preserve critical information (e.g., subject, object) on TACRED while the same quality control cannot be done on MNLI. This further shows the need for a better text simplification control method for MNLI. 

\begin{table}[]
\scalebox{1.02}{%
\begin{tabular}{l l l}
\toprule
 \textbf{Gold Data} & \textbf{Our Approach} & \textbf{Baseline} \\
\midrule
 \textbf{P:} In the apt description of one   & \textbf{P:} It drops below the radar  & \textbf{Predict:} \\
 witness, it drops below the radar & screen ... you don't know & \\
 screen ... you don't know where it is.   & where it is. & Neutral \\
 \textbf{H:} It is hard & \textbf{H:} It is hard for one to & \\ 
 for one to realize  what just happened.  &  find what just happened. & \\
    \textbf{Gold Label:} Entailment  & \textbf{Predict:} Entailment  & \\ \\
\midrule
 \textbf{P:} The tourist industry continued to   & \textbf{P:} The tourist industry  & \textbf{Predict:} \\
   expand, and though it & continued to expand, and & \\
   the top two income earners in & ... top two income earners in & Contradict \\
   Spain, was ... consequences & Spain. ... consequences. & \\
   \textbf{H:} Tourism is not very big & \textbf{H:} Tourism is very big in & \\
   in Spain. & Spain. & \\ 
   \textbf{Gold Label:} Contradict & \textbf{Predict:} Entailment & \\ \\
\midrule
\textbf{P:} This site includes a  & \textbf{P:} This site includes a  & \textbf{Predict:} \\
 list of all award winners and a & list of all award winners and a & \\
 searchable database of Government & searchable database of  & Neutral \\
 Executive articles. & Government Executive articles. & \\
 \textbf{H:} The Government Executive  & \textbf{H:} The Government Executive  & \\
 articles housed on the website & articles are not able to & \\
 are not able to be searched.  &find the website to be searched. &  \\
 \textbf{Gold Label:} Contradict & \textbf{Predict:} Contradict & \\ \\
\bottomrule
\end{tabular}}
\caption[Qualitative comparison of the outputs from our approach (text simplification by ACCESS) and the respective BERT baseline on the original MNLI data. \emph{P, H} indicates premise and hypothesis.]{\doublespacing Qualitative comparison of the outputs from our approach (text simplification by ACCESS) and the respective BERT baseline on the original MNLI data. \emph{P, H} indicates premise and hypothesis.}
\label{tab:example_mnli}
\end{table}

\newpage
On the other hand, table \ref{tab:chapter4_results_mnli_qc} shows results when applying the text simplification control method at inference time mentioned in section \ref{sec:input_simplification}. There is performance improvement when the model is tested on good simplified input (an absolute increase of 1.23\% - 1.18\%, subtracting two columns in row 1). When testing the neural model trained on additional data augmented through a TS method, we further achieve an increase of 0.94\% -- 1.02 \% in model performance (subtract row 2, column 2 with row 2, column 2 in table \ref{tab:chapter4_results_mnli_qc}), showing that we can synthesize two usages of TS in assisting the underlying neural model. All in all, the results show that with a proper quality control method, text simplification can be used to advance neural models' ability to comprehend underlying information encoded in the input text, thus improving the model's overall performance in a downstream natural language task.

\begin{table}[]
    \centering
    \begin{tabular}{l l c c}
        \toprule
        & \textbf{Model} & \textbf{Original Test Data} & \textbf{ Neural TexTSQualiC} \\
        &  & m/mm acc & m/mm acc \\
        \midrule
        1 & BERT fine-tuned on & 82.89/83.10 & 84.12/84.28 \\
          & original training data &  & \\ \\
        \midrule
        2 & BERT fine-tuned on & 83.47/83.61 &  \textbf{84.41}/\textbf{84.63} \\
          & original + simplified & & \\
          & training data & & \\
        \bottomrule
    \end{tabular}
    \caption[Matched (m) and mismatched (mm) accuracies on MNLI development set using 2 BERT-based models: (1) BERT finetuned on original training data, i.e., no training data was simplified, and (2) BERT finetuned on a larger training set that combines original and simplified training data.]{Matched (m) and mismatched (mm) accuracies on MNLI development set using 2 BERT-based models: (1) BERT finetuned on original training data, i.e., no training data was simplified (row 1) and (2) BERT finetuned on a larger training set that combines original and simplified training data (row 2). ``Neural TextTSQualiC'' indicates the model is tested using input selected from a neural classifier that controls simplification quality. Text in bold indicates the best model performance.}
    \label{tab:chapter4_results_mnli_qc}
\end{table}

\section{Conclusion \label{sec:chapter4_conclusion}}

In this chapter, we investigated the effects of neural TS systems on downstream NLP tasks using two strategies: (a) augmenting data to provide machines with additional information during training, and (b) augmenting data to provide machines with additional information during training. Our experiments indicate that the former strategy consistently helps multiple NLP tasks, regardless of the underlying method used to address the task, or the neural approach used for text simplification. On input simplification, we found that, with a good simplification quality control method, text simplification can be used to reduce text difficulty for neural methods. Additionally, this contributes towards eliminating the need for expensive retraining often required by large pretrained neural language models. In this chapter, we have presented our second solutions to data scarcity problem and evaluated its effectiveness across multiple natural language tasks.

\chapter{Neural Ensemble Approach as a Solution for Low-resource \\ Medical Autocomplete Text Simplification \label{chapter:neural_ensembel_approach}} 
\thispagestyle{fancy} 

In previous chapters, we have explored two solutions for data augmentation that work for a variety of NLP tasks. In chapter \ref{chapter:aug_data_qa}, we examined how additional training data augmented by moving answer spans helps improve document retrieval performance in a variety of question-answering systems. Further, in chapter \ref{chapter:aug_data_ts}, we introduced the effective use of text simplification in mitigating the scarcity of data for neural network methods. Specifically, we used text simplification in augmenting training data for neural network methods. With appropriate control of simplification quality, we showed that text simplification methods can be used to help reduce textual complexity, thus enhancing the model's learning process of downstream NLP tasks. Empirically, we have shown that the proposed data augmentation techniques are simple but effective in improving neural network performances. In a nutshell, the two proposed solutions analyzed what kind of information a neural model needs to complete a prediction task and used them to mitigate limitations introduced due to data scarcity for a given task. In this last chapter, we continue to analyze how data scarcity affects neural network models in the learning process and use it to advance the state-of-the-art on a low-resource medical text simplification (MTS) task via a neural ensemble approach.

\begin{figure}
    \centering
    \includegraphics[scale=0.47]{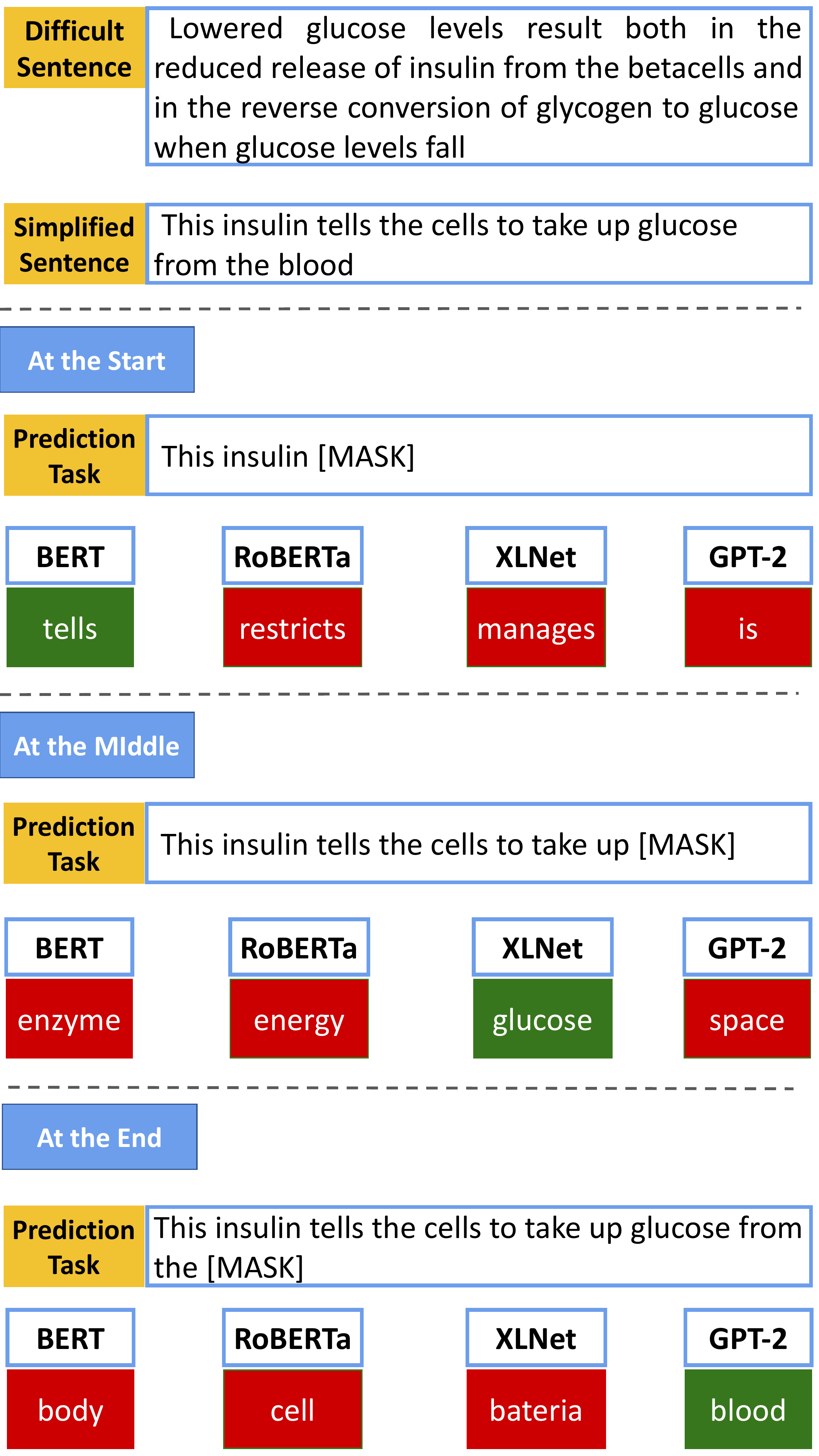}
    \caption[An example of how individual PLMs handle a given medical text simplification task at different stages of the task, e.g. at the start (top), in the middle (middle), and at the end (bottom). Here, we are evaluating 4 PLMs: RoBERTa, BERT, XLNet, and GPT-2.]{An example of how individual PLMs handle a given medical text simplification task at different stages of the task, e.g. at the start (top), in the middle (middle), and at the end (bottom). Here, we are evaluating 4 PLMs: RoBERTa, BERT, XLNet, and GPT-2. Boxes with green color indicate correct predictions and red for incorrect ones. We can see that BERT can predict well at the start of the generation while XLNet excels in the middle, and GPT-2 correctly handle at the end.}
    \label{fig:visualization_example_of_approach}
\end{figure}

The key intuition in this chapter is that despite core architectural designs shared among pretrained language models (PLMs), there are significant differences among pretrained neural language models in their training objectives, thus resulting in a wide spectrum of performances. In section \ref{section:empirical_results_5}, we show that an individual PLM handles inputs varying in characteristics, e.g. sentence length, differently. For example, Figure \ref{fig:visualization_example_of_approach} shows a visualization of how individual PLMs simplify a given difficult sentence, \textit{ ``Lowered glucose levels result both in the
reduced release of insulin from the beta cells and
in the reverse conversion of glycogen to glucose
when glucose levels fall" }, at different time stamps, e.g. beginning, middle, and the end of the sequence generation. The goal of the task is to generate a simplified sentence \textit{This insulin tells the cells to take up glucose
from the blood}. Here, we are evaluating 4 PLMs: RoBERTa, BERT, XLNet, and GPT-2. Boxes with green color
indicate correct predictions and red for incorrect ones. As shown, BERT can predict well at
the start of the generation while XLNet excels in the middle and GPT-2 correctly handles at the end. This partial learning of a problem has been shown to be more severe in a low-resource setting where annotated data is limited \citep{shorten2019survey}. Inspired by this analysis, we propose a simple neural ensemble approach that constitutes partial knowledge learned from individual PLMs, thus creating a good learner of a low-resource MTS task. 

\newpage
In particular, our contributions are as follows:

\textbf{(1)} We explore the application of pretrained neural language models to the autocomplete process for sentence-level medical text simplification. Specifically, given (a) a difficult sentence a user is trying to simplify and (b) the simplification typed so far, the goal is to correctly suggest the next simple word to follow what has been typed. Table \ref{tab:chapter5_example} shows an example of a difficult sentence along with a simplification that the user has partially typed. An autocomplete model predicts the next word to assist in finishing the simplification, in this case, a verb like ``take'', which might be continued to a partial simplification of ``take up glucose''. By suggesting the next word, the autocomplete models provide appropriate guidance while giving full control to human experts in simplifying text.  We explore this task in the health and medical domain where information preservation is a necessity.

\textbf{(2)} We introduce a new medical parallel English corpus of 3,300 sentence pairs for medical text simplification. Each pair consists of a medical difficult sentence and its simplified corresponding version. Our corpus is significantly larger by 45\%  (2,267 pairs vs. 3,300 pairs) in comparison with \citep{van2019evaluating} and uses a stricter
criteria for identifying medical sentence pairs. In particular, \citeauthor{van2019evaluating} only required a single word match in the text itself (vs. ours requires both in the text and the title)
and used a lower similarity threshold of 0.75 (vs. our approach of 0.85).

\textbf{(3)} We also propose a simple but effective ensemble approach that constitutes partial knowledge learned by the four PLMs and outperforms the best individual model by 2.1\%, resulting in an overall word prediction accuracy of 64.52\% on the new medical corpus.

\section{Chapter Overview}

The remainder of the chapter is organized as follows. We discuss the background work ins medical text simplification in the next section. The new medical parallel English corpus for MTS is presented in section \ref{sec:chapter5_datasets}. Our proposed ensemble approach for autocomplete sentence-level medical text simplification is explained in section \ref{sec:chapter5_approach}. Section \ref{sec:chapter5_implementation} provides information on our implementation details. The empirical results and discussion of our proposed approach are detailed in section \ref{section:empirical_results_5}. Finally, section \ref{sec:chapter5_conclusion} summarizes the key conclusions of the work presented in this chapter.

\section{Background Work}

The goal of text simplification is to transform text into a variant that is more broadly accessible to a wide variety of readers while preserving the content. Plain language in medicine has long been advocated as a way to improve patient understanding and engagement \citep{ondov2022survey}. To this end, in recent years, there has been a growing effort in simplifying medical text and creating more patient-friendly and -accessible documents. Advances in machine translation have also been applied to simplify medical documents \citep{wang2016text}. Following this approach, there is an increased interest in improving sequence-to-sequence model performance on reducing the textual complexity of medical documents, making them more accessible to patients with zero to little knowledge of medicine \citep{pattisapu2020leveraging,chen2017automatic,qenam2017text,luo2022benchmarking}. There are also successful attempts at using lexical text simplification techniques to improve health literacy for patients. For example, \citep{leroy2013user} used term familiarity to simplify medical text and evaluate patient understanding of medical documents. Similarly, there has been efforts in extending the use of phrase tables in simplifying medical lexicons while preserving the sentence structure \citep{abrahamsson2014medical,shardlow2019neural,swain2019lexical}. Further, recognizing the significantly enhanced outcomes of patient-doctor communication when using plain language, researchers have been exploring text simplification in languages other than English \citep{cardon2020french,vstajner2015automatic,klaper2013building}.

However, it has been shown that, in some domains, such as healthcare, using fully-automated text simplification is not appropriate since it is critical that the important information is preserved fully during the simplification process \citep{begoli2019need}. For example, \citep{shardlow2019neural} found that fully automated approaches omitted 30\% of critical information when used to simplify clinical texts. Further, in 65.9\% majority of the time, the sequence-to-sequence TS system does not make changes to the text, proving that fully-automated systems are incapable of identifying difficult sentences to simplify. Therefore, for these very specialized domains, instead of fully-automated approaches, interactive text simplification tools are better suited to generate more efficient and higher quality simplifications \citep{kloehn2018jmir}.

Autocomplete tools suggest one or more words during text composition that could follow what has been typed so far. They have been used in a range of applications including web queries \citep{cai2016survey}, database queries \citep{khoussainova2010snipsuggest}, texting \citep{dunlop2000predictive}, e-mail composition \citep{chen2019gmail}, and interactive machine translation,  where a user translating a foreign sentence is given guidance as they type \citep{green-etal-2014-human}. Our work is most similar to interactive machine translation. 
Autocomplete tools can speed up the text simplification process and give full control over information preservation to users, which is required in some domains, such as health and medicine.

\section{Datasets \label{sec:chapter5_datasets}}

\begin{sidewaystable}
    \centering
    \begin{tabular}{l l l}
    \toprule
    & \textbf{Difficult} & \textbf{Simple} \\
    \midrule
    1 & \textbf{Lowered glucose levels result both in the reduced release} & \textbf{This insulin tells the cells to take up glucose from the }\\
      & \textbf{of insulin from the beta cells and in the reverse} & \textbf{blood.} \\
      & \textbf{conversion of glycogen to glucose when glucose levels fall.} & \\
    \midrule
     2 & A 2003 study found that regular use of chopsticks by the & It found that people who use chopsticks regularly have\\
       &  elderly may slightly increase the risk of  osteoarthritis in the &  a slightly higher risk of getting arthritis in the hand .\\
       &  hand , a condition in which cartilage is &  With this , cartilage gets worn off , which causes pain\\
       &  worn out , leading to pain and swelling in the hand joints. & in the joints of the hand .\\
    \midrule
    3 & Leptospirosis -LRB- also known as Weil 's disease , Weil 's & Leptospirosis -LRB- also known as Weil 's disease , \\
      & syndrome , canicola fever , canefield fever , nanukayami & canicola fever , canefield fever , nanukayami fever or \\
      & fever , 7-day fever , Rat Catcher 's Yellows , Fort Bragg & seven day fever -RRB- is a bacterial disease . It is \\
      & fever , and Pretibial fever -RRB- is a bacterial zoonotic & caused by spirochaetes of the genus Leptospira . \\
      & disease caused by spirochaetes of the genus Leptospira that & \\
      & affects humans and a wide range of animals , including & \\
      & mammals , birds , amphibians , and reptiles . & \\
    \midrule
    4 & Spontaneous abortion -LRB- also known as miscarriage -RRB- & People speak of spontaneous abortion or miscarriage  \\
      & is the expulsion of an embryo or fetus due to accidental  & when the embryo or fetus is lost due to natural causes \\
      & trauma or natural causes before approximately the 22nd week &  before the 20th week of pregnancy . \\
      & of gestation ; the definition by gestational age. & \\
    \midrule
    5 & Aneurysms can commonly occur in arteries at the base of the  & Aneurysms usually happen in arteries at the base of the \\
      & brain -LRB- the circle of Willis -RRB- and an aortic & brain and in the aorta -LRB- the main artery coming \\
      & aneurysm occurs in the main artery carrying blood from the & out of the heart -RRB- - this is an aortic aneurysm .\\
      & left ventricle of the heart . & \\
    \bottomrule
    \end{tabular}
    \caption[Examples of sentence pairs in Medical Wikipedia parallel corpus. The sentence in bold is used as a walk-through example in the remainder of this chapter.]{Examples of sentence pairs in Medical Wikipedia parallel corpus. The sentence in bold is used as a walk-through example in the remainder of this chapter.}
    \label{tab:medexample}
\end{sidewaystable}

To evaluate our approach, we curated a dataset for medical text simplification as follows. We automatically extracted medical sentence pairs from the sentence-aligned English Wikipedia corpus \citep{kauchak2013improving}. Table \ref{tab:medexample} shows a set of medical sentence pairs. To identify the medical sentence pairs, we first created a medical dictionary with 260k medical terms selected from the Unified Medical Language System (UMLS) \citep{bodenreider2004unified} by selecting all UMLS terms that were associated with the semantic types of: Disease or Syndrome, Clinical Drug, Diagnostic Procedure, and Therapeutic or Preventive Procedure. The UMLS is a set of files and software that brings together many health and biomedical vocabularies and standards to enable interoperability between computer systems. The UMLS includes key biomedical terminology, classification and coding standards, and associated resources. There are three knowledge sources within the UMLS system: Metathesaurus, Semantic Networks, Specialist Lexicon, and Lexical Tools. In this work, we utilized a Metathesaurus knowledge source to extract biomedical vocabulary.

We extracted sentences from the English Wikipedia corpus based on the occurrence of terms in the medical dictionary.  Specifically, a sentence pair was identified as medical
if both the title of the article and the English Wikipedia sentence had 4 or more terms that matched the medical keywords. A term was considered a match to the UMLS dictionary if it had a similarity score higher than 0.85 using QuickUMLS \citep{soldaini2016quickumls}. The final medical parallel corpus has 3.3k aligned sentence pairs\footnote{\url{https://github.com/vanh17/MedTextSimplifier/tree/master/data_processing/data}}.

Table \ref{tab:medsize} shows the corpus size for the two corpora.

\begin{table}
    \centering
    \scalebox{1.0}{%
    \begin{tabular}{l c}
    \toprule
    \textbf{Domain}  & \textbf{No. Sentence Pairs} \\
    \midrule
    General Domain & 163,700 \\
    Medical Domain & 3,300 \\
    \midrule
    Total & 167,000\\
    \bottomrule
    \end{tabular}}
    \caption[Number of sentence pairs for General Domain and Medical Domain. Note that the two corpora are exclusive. These sentence pairs are identified as medical if they contain four or more medical terms.]{Number of sentence pairs for General Domain and Medical Domain. Note that the two corpora are exclusive. These sentence pairs are identified as medical if they contain four or more medical terms. The medical terms used for the above process are extracted from multiple semantic types, such as of Disease, Syndrome, Clinical Drug, Therapeutic, and Preventive Procedure in the Unified Medical Language System (UMLS).}
    \label{tab:medsize}
\end{table}

Simultaneously, there is also a parallel medical corpus created by filtering sentence pairs from Wikipedia \citep{van2019evaluating}.  Our corpus is significantly larger (45\% larger; 2,267 pairs vs. 3,300 pairs) and uses stricter criteria for identifying sentences: they only required a single word match in the text itself (not the title) and used a lower similarity threshold of 0.75 (vs. our approach of 0.85). We used stricter criteria to ensure non-medical sentence pairs were included in the final corpus. However, our approach results in an additional 1033 sentence pairs. Note that to be extracted, sentence pairs in our medical corpus need to include both a medical title and at least 4 medical terms as opposed to at least 1 match and with no requirement on document title in \citep{van2019evaluating}. We also used higher similarity threshold for matching words in simple sentence with UMLS terms ($0.85$ vs $0.75$). Table \ref{tab:medicalwar} provides a comparison between our medical corpus and the one created in \citep{van2019evaluating}.

\begin{table}
     \centering
     \scalebox{1.0}{%
     \begin{tabular}{l c}
    \toprule
     Corpus  & No. Sentence Pairs \\
     \midrule
     Medical Parallel English Wikipedia Corpus & \textbf{3,300} \\
     Van den Bercken et al., 2019 & 2,267 \\
     \bottomrule
     \end{tabular}}
     \caption[A comparison between the Medical Parallel English Wikipedia Corpus and \citep{van2019evaluating}. Note that to be extracted, sentence pairs in our medical corpus need to include both medical title and at least 4 medical terms as opposed to at least 1 match and with no requirement on document title in \citep{van2019evaluating}.]{A comparison between the Medical Parallel English Wikipedia Corpus and \citep{van2019evaluating}. Note that to be extracted, sentence pairs in our medical corpus need to include both medical title and at least 4 medical terms as opposed to at least 1 match and with no requirement on document title in \citep{van2019evaluating}.}
     \label{tab:medicalwar}
     \
\end{table}

\section{Approach \label{sec:chapter5_approach}}

We had three goals for our analysis: explore a new application for pretrained language models to autocomplete medical text simplification, understand the effect of incorporating the additional context of the difficult sentence into autocomplete TS models, and evaluate our new ensemble approach to the autocomplete TS. 

\subsection{Autocomplete Approach For Medical Text Simplification}
We pose the autocomplete text simplification problem as a language modeling problem: given a difficult sentence that a user is trying to simplify, $d_1 d_2 ... d_m$, and the simplification typed so far, $s_1 s_2 ... s_i$, the autocomplete task is to suggest word $s_{i+1}$.  Table \ref{tab:chapter5_example} gives an example of the autocomplete task. To evaluate the models, we calculated how well the models predicted the next word in a test sentence, given the previous words.  A simple test sentence of length $n$, $s_1 s_2 ... s_n$, would result in $n-1$ predictions, i.e., predict $s_2$ given $s_1$, predict $s_3$ given $s_1 s_2$. Algorithm \ref{alg:chapter5_autocomplete} details the complete process for autocomplete text simplification. Here, $next\_word\_prediction$ is a masked language model trained using the following loss function:
\begin{equation}
    \resizebox{0.6\hsize}{!}{$L_{mlm}(x|s_{1}s_{2}s_{3}...s_{i}) = \log p(x|s_{1}...s_{i};\theta)$}
\end{equation}

\noindent Specifically, Table \ref{tab:medexample} shows a difficult sentence from medical domain and the corresponding simplification from the same domain. Given this test example, we generate 12 prediction tasks, one for each word in the simple sentence after the first word.  Table \ref{fig:testing} shows these 12 test prediction tasks.

\begin{table}
    \centering
    \scalebox{1.03}{%
    \begin{tabular}{l l}
    \toprule
    Difficult & Lowered glucose levels result both in the reduced release of insulin from \\
              & the beta cells and in the reverse conversion of glycogen to glucose when \\ 
              & glucose levels fall. \\
    \hline
    Typed &  This insulin tells the cells to $\rule{1cm}{0.15mm}$ \rule{0pt}{3ex} \\
    \bottomrule
    \end{tabular}}
    \caption[An example of text simplification Autocomplete task.  The user is simplifying the difficult sentence on top and has typed the words on the bottom so far. The example is taken from a medical parallel English Wikipedia sentence pair in Table \ref{tab:medexample}.]{ An example of text simplification Autocomplete task.  The user is simplifying the difficult sentence on top and has typed the words on the bottom so far. The example is taken from a medical parallel English Wikipedia sentence pair in Table \ref{tab:medexample}.}
    \label{tab:chapter5_example}
\end{table}

\begin{algorithm}[hbt!]
\caption{Autocomplete Text Simplification Algorithm}\label{alg:chapter5_autocomplete}
\KwData {difficult sentence: $ d_1 d_2 ... d_m $} 

\KwResult{simple sentence: $s_1 s_2 ... s_n$}

$curr\_sim \gets s_1 [masked]$\;
$masked \gets next\_word\_prediction(curr\_sim)$;

\While{$masked \neq [end]$}{

    {$curr\_sim \gets curr\_sim + masked$};
    
    {$masked \gets next\_word\_prediction(curr\_sim)$};
}
\end{algorithm}

We examined four pretrained neural language models based on the Transformer network \citep{vaswani2017attention}: BERT \citep{devlin2018bert},  RoBERTa \citep{liu2019roberta},  XLNet \citep{yang2019xlnet},  and GPT-2 \citep{radford2019language}.  For each of the models, we examine versions that only utilize the text typed so far, denoted ``No Context'', as well as ``Context-Aware'' variants that utilize the difficult sentence. Specifically, for the ``No Context'' variants, we predict the next word for the input ``$s_1 s_2 ... s_i \mbox{[NEXT]} .$''. Unlike other autocomplete tasks, for text simplification, the difficult sentence provides very explicit guidance about what words and information should be included as the user types. We take a straightforward approach to incorporate the context of the difficult sentence: we concatenate the difficult sentence in front of a simplification typed so far, i.e., predict $s_{i+1}$ given $d_1 d_2 ... d_m [sep] s_1 s_2 ... s_i$.  This has the advantage of biasing the predictions towards those found in the encoded context from difficult sentences, but is still model-agnostic, allowing us to apply it to all the different pretrained languge models without modifying the underlying architecture. 

We fine-tuned all four models on the 160k sentence pair general parallel English Wikipedia \citep{kauchak2013improving} (excluding the development and testing data) and then further fine-tuned them on the separate medical training set described in section \ref{sec:chapter5_datasets}.
Note that none of the test sentences was in a dataset used for fine-tuning.

\paragraph{BERT:} Bidirectional Encoder Representations from Transformers  \citep{devlin2018bert} is a method for learning language representations using bidirectional training. BERT has been shown to produce state-of-the-art results in a wide range of generation and classification applications. We use the base original BERT\footnote{\url{https://github.com/huggingface/bert/}} model pre-trained on the BooksCorpus \citep{zhu2015aligning} and English Wikipedia.

\paragraph{RoBERTa:} A Robustly Optimized BERT Pretraining Approach \citep{liu2019roberta}. The RoBERTa uses the same model architecture as BERT. However, the differences between RoBERTa and BERT are that RoBERTa does not use Next Sentence Prediction during pre-training and RoBERTa uses larger mini-batch size. We used the publicly released base RoBERTa\footnote{\url{https://github.com/huggingface/roberta}} with 125M parameters model.

\paragraph{XLNet:} Generalized Auto-regressive Pretraining Method \citep{yang2019xlnet}. Like BERT, XLNet benefits from bidirectional contexts. However, XLNet does not suffer limitations of BERT because of its auto-regressive formulation. In this work, we used publicly available base English XLNet\footnote{\url{https://github.com/huggingface/xlnet}} with 110M parameters model.

\paragraph{GPT-2:} Generative Pretrained Transformer 2 \citep{radford2019language}. Like BERT, GPT-2 is also based on the Transformer network, however, GPT-2 uses unidirectional left-to-right pre-training process. We use the publicly released GPT-2\footnote{\url{https://github.com/huggingface/gpt2}} model, which has 117M parameters and is trained on web text.

\begin{table}
\centering
\scalebox{.89}{%
\begin{tabular}{c l l}
\toprule
Difficult sentence & Lowered glucose levels result both in the reduced release of insulin & \\
 & from the betacells and in the reverse conversion of glycogen to   & \\
 & glucose when glucose levels fall. & \\
\midrule
\toprule
Prediction Task & Simplification typed so far & Predict \\
\midrule
1 & This & insulin \\
2 & This insulin & tells \\
3 & This insulin tells & the \\
4 & This insulin tells the & cells \\
5 & This insulin tells the cells & to \\
6 & This insulin tells the cells to & take \\
7 & This insulin tells the cells to take & up \\
8 & This insulin tells the cells to take up & glucose \\
9 & This insulin tells the cells to take up glucose & from \\
10 & This insulin tells the cells to take up glucose from & the \\
11 & This insulin tells the cells to take up glucose from the & blood \\
12 & This insulin tells the cells to take up glucose from the blood & .\\
\bottomrule
\end{tabular}}
\caption{The resulting prediction tasks that are generated from the example in Table \ref{tab:medexample}.}
\label{fig:testing}
\end{table}

\subsection{Ensemble Models} \label{sec:ensemble}

Each of the models above utilizes different network variations and was pretrained on different datasets and objectives. Therefore, they do not always make the same predictions.  Ensemble approaches combine the output of different systems to try and leverage these differences to create a single model that outperforms any of the individual models. Specifically, Algorithm \ref{alg:chapter5_ensemble} explains the general architecture of ensemble approaches. The key to successfully ensembling output from a variety of models lies in choosing a correct scoring function. In this chapter, we examined three different scoring functions for ensemble approaches that combine the output of the four pretrained language models.

\begin{algorithm}[hbt!]
\caption{Neural Ensemble Approach for Autocomplete Text Simplification}\label{alg:chapter5_ensemble}
\KwData {predictions from MLMs: [$pred_1, pred_2, ..., pred_m $]};
        {input: $X = d_1 d_2 d_3 ... [sep] s_1 s_2 ... [masked]$};

\KwResult{: the best prediction across all MLMs }

$highest\_score = scoring\_function(X, pred_1)$;

$best\_pred = pred_1$;

\While{ pred in $[pred_1, pred_2, pred_3, ..., pred_m]$ }{
    { $curr\_score \gets scoring\_function(X, pred)$ };
    
    \If { $curr\_score < highest\_score$ } {
        {$highest\_score \gets curr\_score$};
        
        {$best\_pred \gets pred$};
    }
}
\end{algorithm}

\newpage

\paragraph{Majority Vote:} As a baseline ensemble approach, we examined a simple majority vote on what the next word should be. We take the top 5 suggestions from each of the models and do a majority count on the pool of combined suggestions.  The output of the model is the suggestion with highest count. If there is a tie, we randomly select one of the top suggestions. We picked the top 5 suggestions since this was the cutoff where the models tended to peak on the development data (for example, see the accuracy@N as shown in Table \ref{tab:accuracy-@-n}. Having more than 5 suggestions did not improve performance much but slowed the computation.

\paragraph{4-Class Classification (4CC):} \label{sec:4cc} The ensemble problem can be viewed as a classification problem where the goal is to predict which system output should be used given a difficult sentence and the words typed so far, i.e., the autocomplete example. We posed this as a supervised classification problem.  Given an autocomplete text simplification example, we can generate training data for the classifier by comparing the output of each system to the correct answer. If a system does get the example correct, then we include an example with that system as the label. Table \ref{tab:4ccexample} shows three such examples, where RoBERTa correctly predicted the first example, BERT the second, and XLNet the third.  If multiple systems get the example right, we then randomly assign the label to one of the systems.

We train a neural text classifier implemented by huggingface\footnote{\url{https://github.com/huggingface/transformers}} with this training set to the pretrained language models given the next-word prediction task. To make use of the models' confidence on top of the results from model selection, we designed a scoring system for output selection as follows:
\begin{equation} \label{eq:4cc}
	\resizebox{.6\hsize}{!}{$Score(w, X) = \alpha * P(w|X) + \theta * I (X, S)$}
\end{equation}

\noindent where $P(w|X)$ is model $X$'s confidence on predicted word $w$; $I(X, S)$ is an identity function, which returns $1$ if $X = S$ and $0$ otherwise; $S$ is the predicted model from model selector; and $\alpha$ and $\theta$ are scoring parameters. We use $0.5$ for both $\alpha$ and $\theta$.  
At testing time, we pick the highest score and output the word $w$, given a prediction task.
\begin{table}
    \centering
    \scalebox{1.0}{%
    \begin{tabular}{l l}
    \toprule
    Prediction Task  & Class \\
    \midrule
    (Difficult sentence). This (MASK) & RoBERTa \\
    (Difficult sentence). This insulin (MASK) & BERT \\
    (Difficult sentence). This insulin tells (MASK) & XLNet \\
    \bottomrule
    \end{tabular}}
    \caption{An example of training data for the 4CC model. Class can be one of the four option: BERT, RoBERTa,  XLNet, GPT-2.}
    \label{tab:4ccexample}
    \
\end{table}

\paragraph{Autocomplete for Medical Text Simplification (AutoMeTS):} 

One potential problem with the 4CC approach is that the training data for the ensemble approach might be biased towards the best individual model. This is mostly because the best performer tends to score better than the majority of the candidates. To mitigate this effect, we also developed an ensemble approach based on a multi-label classifier for model selector, which we denote the AutoMeTS ensemble model. This choice of model selector, to our knowledge, is novel to transformer-based ensemble models. For this choice of classifier, each prediction task is given a sequence of 4 binary labels. Each label represents the correctness of each of the individual pretrained language models, with a 0 representing an incorrect prediction on the task and a 1 representing a correct prediction. Table \ref{tab:AutoMeTSexample} shows an example of this dataset with the labels in order ``RoBERTa BERT XLNet GPT-2''.  For the first example, RoBERTa, XNLET, and GPT-2 correctly predicted the next word, while BERT did not.

We trained a neural multi-label classifier implemented by huggingface on this training dataset. To make use of the models' confidence on top of the results from model selection, we designed a scoring system for output selection as follows:

\begin{equation}
    \label{eq:AutoMeTS}
	\resizebox{0.6\hsize}{!}{$Score(w, X) = \beta * P(w|X) + \sigma * S(X, Ls)$}
\end{equation}

\noindent where $P(w|X)$ is model $X$'s confidence on predicted word $w$; $S(X, Ls)$ is a function, which returns $0.25$ if model $X$ is in $Ls$ and $0$ otherwise; $Ls$ is the predicted sequence of labels from the model selector; and $\beta$ and $\sigma$ are scoring parameters. We use $0.5$ for both $\beta$ and $\sigma$. At testing time, we output the word $w$ with the highest score, given a prediction task.

\begin{table}
    \centering
    \scalebox{1.0}{%
    \begin{tabular}{l c}
    \toprule
    Prediction Task  & Sequence of Labels \\
    \midrule
    (Difficult sentence). This (MASK) & 1 0 1 1 \\
    (Difficult sentence). This insulin (MASK) & 0 1 0 0 \\
    (Difficult sentence). This insulin tells (MASK) & 1 1 1 1 \\
    \bottomrule
    \end{tabular}}
    \caption{An example of training data for the AutoMeTS model. For a prediction task, a sequence of 4 labels is give in the order "RoBERTa BERT XLNet GPT-2". The value of 1 means the model correctly predicted the right word, and 0 otherwise.}
    \label{tab:AutoMeTSexample}
    \
\end{table}

\section{Implementation Details \label{sec:chapter5_implementation}}

We compare the performance of the models on the medical autocomplete text simplification task.  We used  our medical parallel English Wikipedia corpus with 70\% of the sentence pairs for training, 15\% for development, and 15\% for testing. We fine-tuned individual pretrained neural language models using huggingface\footnote{\url{https://github.com/huggingface/}} with a batch-size of 8, 8 epochs, and a learning rate of $5e^{-5}$. Early stopping was used based on the second time a decrease in the accuracy was seen.
\paragraph{No-context:} We calculated how well each model predicts the next word in a test sentence, given the previous words.  A test sentence of length $n$, $s_1 s_2 ... s_n$, would result in $n-1$ predictions, i.e., predict $s_2$ given $s_1$, predict $s_3$ given $s_1 s_2$, etc.  For example, Table \ref{tab:medexample} shows a difficult sentence from English Wikipedia and the corresponding simplification from the medical Simple English Wikipedia. For this test example, we generate 12 prediction tasks, one for each word in the simple sentence after the first word.  Table \ref{fig:testing} shows these 12 test prediction tasks.  
\paragraph{Context-aware:} For the context-aware approaches, a corresponding difficult sentence is concatenated as a prefix for each prediction task. We measured the performance of a system using accuracy based on the number of predictions that exactly matched the next word in the corpus.  

\paragraph{Evaluation Measures:} We used two metrics to evaluate the quality of the approaches.  First, we used standard accuracy, where a prediction is counted correct if it matches the test prediction word. Accuracy is pessimistic in that the predicted word must match exactly the word seen in the simple sentence, and as such it does not account for other possible words, such as synonyms, that could be correctly used in the context. Since the parallel English Wikipedia corpus does not offer multiple simplified versions for a given difficult sentence, accuracy is the best metric that considers automated scoring, simplification quality, and information preservation. Accuracy-based metrics can help offset an expensive manual evaluation while providing the best approximation of how the autocomplete systems work. We do not use BLEU \citep{papineni2002bleu} and SARI \citep{xu2016optimizing} scores, which are oftenly used in the text simplification domain, because the two metrics are specifically designed for fully-automated models that predict an entire sentence at a time.  For autocomplete, the models only predict a single word at a time and then, regardless of whether the answer is correct or not, use the additional context of the word that the user typed next to make the next prediction. 

Autocomplete models can suggest just the next word, or they can be used to suggest a list of alternative words that the user could select from (since the models are probabilistic they can return a ranked list of suggestions).  To evaluate this use case, and to better understand what words the models are predicting, we also evaluated the models using accuracy@N. Accuracy@N counts a model as correct for an example as long as it suggests the correct word within the first $N$ suggestions.

\section{Empirical Results and Discussion \label{section:empirical_results_5}}

We first analyze the results of the individual pretrained language models on the medical text simplification autocompletion task and then explore the ensemble approaches.  We also include a number of post-hoc analyses to better understand what the different models are doing and limitations of the models.
\begin{table}[t]
\centering
\scalebox{1.0}{%
\begin{tabular}{l c c }
\toprule
Model & No Context & Context-Aware\\
\midrule
Single PNLMs & & \\
\midrule
RoBERTa & \textbf{56.23} & \textbf{62.40} \\
BERT & 50.43 & 53.28 \\
XLNet & 45.70 & 46.20 \\
GPT-2 & 23.20 & 49.00 \\
\midrule
Ensemble Models & & \\
\midrule
Majority Vote & 39.75 & 48.25 \\
4CC & 52.27 & 59.32 \\
AutoMeTS & \textbf{57.89} & \textbf{64.52} \\
Upper bound & 60.22 & 66.44 \\
\bottomrule
\end{tabular}}
\caption[Accuracy of pretrained neural language models (PNLMs), both with and without the context of the difficult sentence on the medical parallel English Wikipedia corpus.]{Accuracy of pretrained neural language models (PNLMs), both with and without the context of the difficult sentence on the medical parallel English Wikipedia corpus. Note that, ``no-context'' refers to input without encoded context of difficult sentences, i.e., we predict word $s_{i+1}$ given only the simplification typed so far, $s_1 s_2 ... s_i$. For context-aware, we take a straightforward approach to incorporate the context of the difficult sentence: we concatenate the difficult sentence in front of a simplification typed so far, i.e., predict $s_{i+1}$ given $d_1 d_2 ... d_m . s_1 s_2 ... s_i$.} 
\label{tab:chapter5_results}
\end{table}

\subsection{Individual Language Models}

\paragraph{Accuracy} Table \ref{tab:chapter5_results} shows the results for the four different variants (RoBERTa, BERT, XLNet, and GPT-2 with and without context). Even without any context, many of the models get every other word correct (accuracy of around 50\%).  RoBERTa scores significantly higher than other individual models (accuracy of 56.23\%) while GPT-2 struggles the most (accuracy of 23.20\%). This uneven performance across individual models introduces a class imbalance problem, where we have disproportionate number of class instances, to the 4CC ensemble approach later. Specifically, there is a strong bias toward RoBERTa in training data for the 4CC ensemble approach, which will be detailed further in Section \ref{sec:chapter5_results_ensemble} 

\begin{table}[t]
\centering
\scalebox{1.0}{%
\begin{tabular}{l c c c c}
\toprule
 & RoBERTa & BERT & XLNet & GPT-2\\
 \midrule
accuracy@1 & 62.40 & 53.28 & 46.20 & 49.00 \\
accuracy@2 & 67.20 & 54.50 & 46.90 & 49.44\\
accuracy@3 & 70.00 & 56.20 & 49.20  & 52.57\\
accuracy@4 & 72.10 & 58.00 & 51.30 & 54.32\\
accuracy@5 & \textbf{73.20} & \textbf{59.40} & \textbf{53.50} & \textbf{56.12}\\
 \bottomrule
\end{tabular}}
\caption{\label{tab:accuracy-@-n} Accuracy@N of the RoBERTa, BERT, XLNet, and GPT-2 with context on next word prediction.}
\end{table}

\paragraph{Accuracy@N} Table \ref{tab:accuracy-@-n} shows the accuracy@N from pretrained neural language models on next word prediction.  By allowing the autocomplete system user the option to pick from a list of options, the correct word is much more likely to be available.  Even just showing three options, results in large improvements, e.g., 7.5\% absolute for RoBERTa.  When five options are available, increases range from  6--10.8\%.

\paragraph{Impact of the encoded context from the difficult sentences:} Table \ref{tab:chapter5_results} shows performances from both individual and ensemble models on the autocomplete medical text simplification task. Across all models, models predict better when provided additional encoded context from difficult sentences. When compared results from ``no-context'' setting with those from ``context-aware'' setting, there are significant increases from increases of 2.85\% - 26\% in performance accuracy. We hypothesize that this is because, different to other autocomplete tasks, for text simplification, the difficult sentence provides very explicit guidance of what words or terms should be included as the user types. It is worth noting that the enhanced performances come from a simple concatenation of the difficult sentence in front of the simplification typed so far with no change to the underlying architecture of each of the models.

\begin{figure}[t]
\center{\includegraphics[scale=0.678]
{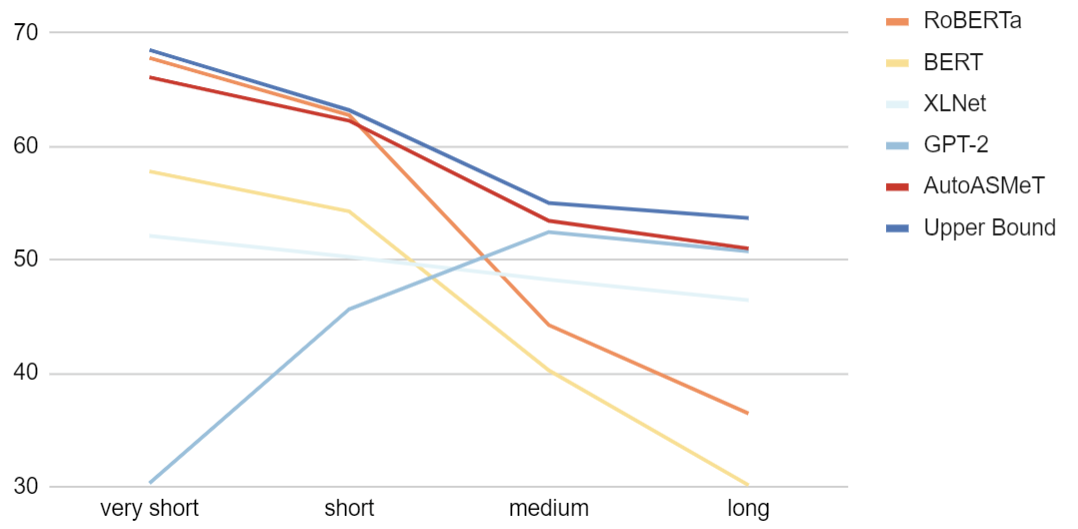}}
\caption[Accuracy for the context-aware models based on the length of the difficult sentences: very short ($\le 5$ tokens), short ($6-15$), medium ($16-19$), and long ($\ge20$).]{\label{fig:length_based} Accuracy for the context-aware models based on the length of the difficult sentences: very short ($\le 5$ tokens), short ($6-15$), medium ($16-19$), and long ($\ge20$).}
\end{figure}

\paragraph{Impact of the difficult sentence length:} To better understand the models, we compared the average performance of the models based on the length of the sentence that was being simplified.  We divided the test sentence into four different groups based on length: very short ($\leq$5 tokens), short ($6-15$ tokens), medium ($16-19$ tokens), and long ($\geq$20 tokens).  Figure \ref{fig:length_based} shows the test accuracy of the context-aware models broken down into these 4 different groups. RoBERTa, BERT, and XLNet are fairly consistent regardless of the difficult sentence lengths; only for long sentences does the performance drop. GPT-2 performs poorly on short sentences, but well for other lengths. We hypothesize that the training data for GPT-2 (web text) may require more context for this more technical task.

\begin{figure}[t]
\center{\includegraphics[scale=0.67]
{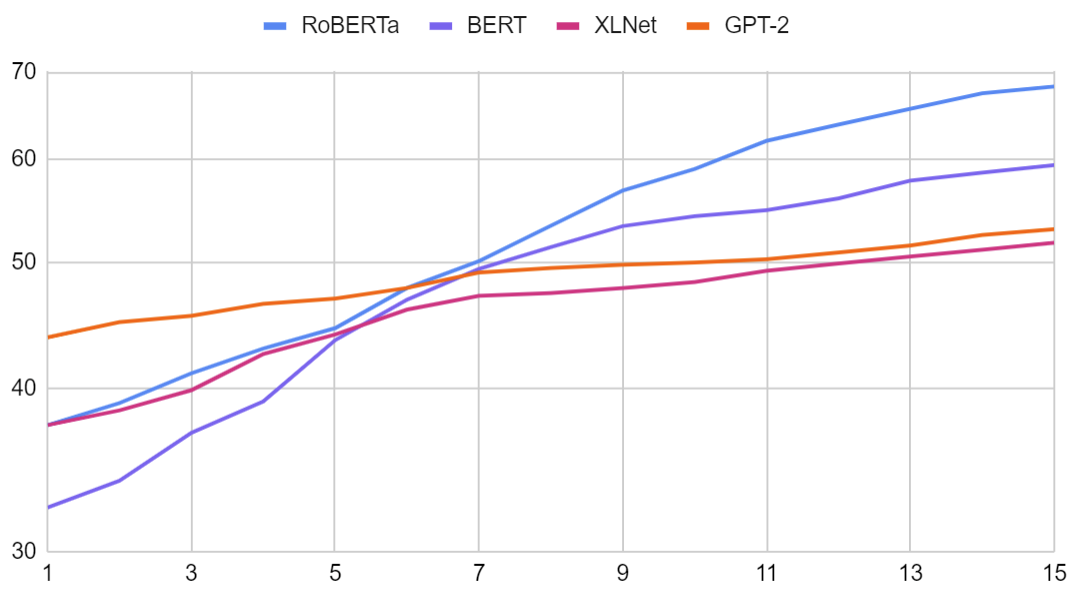}}
\caption[Accuracy for the four context-aware models based on the number of words typed so far.]{Accuracy for the four context-aware models based on the number of words typed so far.}
\label{fig:position_based}
\end{figure}

\paragraph{Impact of the number of words typed:}  We also analyzed the average performance of the models based on how many of the words of the simplified sentence had been typed so far, i.e., the length of $s_1 s_2 ... s_i$.  Figure \ref{fig:position_based} shows the performance accuracy of the models based on how many words of the simplification the model has access to. Early on, when the sentence is first being constructed, all models perform poorly. As more words are typed, the accuracy of all models increases.  GPT-2 performs the best early on, but then levels off after about 7 words.  Both BERT and RoBERTa continue to improve as more context is added, which may partially explain their better performance overall.

\begin{figure}[t]
\center{\includegraphics[scale=.67]
{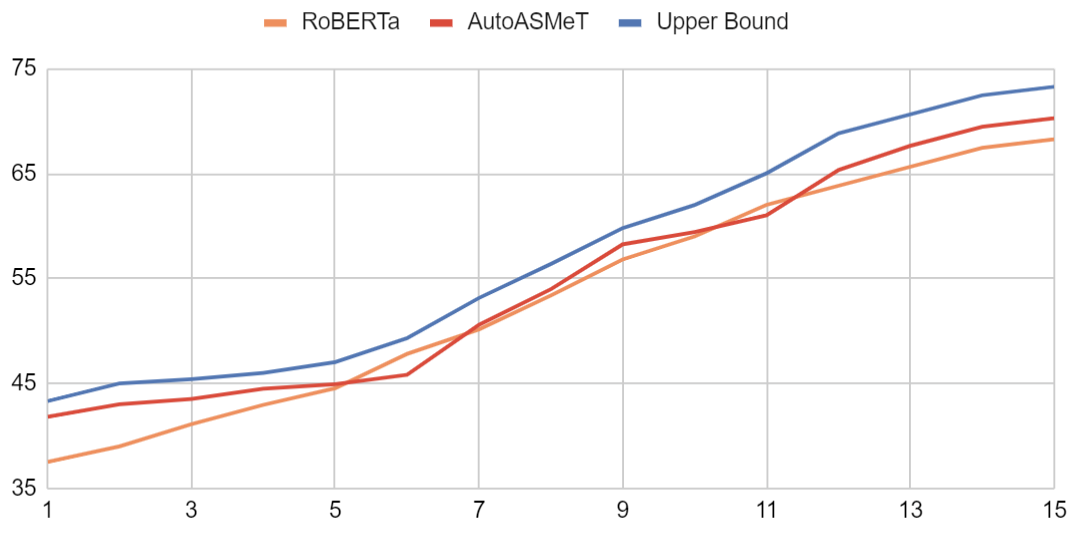}}
\caption[Accuracy for the RoBERTa, AutoMeTS, and upper bound models based on the number of words typed so far.]{\label{fig:ensemble_pos} Accuracy for the RoBERTa, AutoMeTS, and upper bound models based on the number of words typed so far.}
\end{figure} 

\subsection{Ensemble Models \label{sec:chapter5_results_ensemble}}

Although RoBERTa performs the best overall, as Figures \ref{fig:length_based} and \ref{fig:position_based} show, different models perform better in different scenarios.  The ensemble models try and leverage these differences to generate a better overall model.

As shown in Table \ref{tab:chapter5_results}, the majority vote ensemble model does not perform better than the best individual pretrained neural language model, RoBERTa.  Even though the 4CC ensemble approach outperforms the majority vote approach by 11.07\% and performs better than three out of the four pretrained neural language models, it still fails to beat RoBERTa.  AutoMeTS, by viewing the problem as a multi-label problem, is able to avoid some of the biases in the training data that 4CC has, resulting in an absolute improvement of 2.1\% over the best individual model (RoBERTa).

To understand the performance differences of the ensemble models,  table \ref{tab:model_frequency} shows the percentage that each of the four pretrained neural language models was used for each of the ensemble approaches.  The problem with the majority vote is that it tends to utilize all of the systems, regardless of their quality.  For example, it shows a high percentage of XLNet, even though its performance was the worst.  Because RoBERTa is the best-performing model, the 4CC approach had a very strong bias towards RoBERTa. The multi-label selector reduces the bias towards using RoBERTa (an 11.25\% decrease in the use of RoBERTa in the ensemble model) and is able to leverage predictions from the other models when appropriate.

\begin{table}[t]
\centering
\scalebox{1.0}{%
\begin{tabular}{l c c c}
\toprule
	 	&	Majority Vote	&	4CC & AutoMeTS\\
\midrule
RoBERTa & 47.29\%  & 	71.00\% 		& 	59.75\% \\
BERT 	& 20.25\%  & 	12.45\% 		& 	18.09\% \\
XLNet 	& 15.41\%  & 	5.72\% 		& 	7.06\% \\
GPT-2 	& 17.05\%  & 	10.83\% 		& 	15.10\% \\
\bottomrule
\end{tabular}}
\caption[The appearance frequency of pretrained neural language models in Majority Vote, 4CC, AutoMeTS (our approach) ensemble models.]{The appearance frequency of pretrained language models in Majority Vote, 4CC, AutoMeTS (our approach) ensemble models.}
\label{tab:model_frequency}
\end{table}

To understand the limits of an ensemble  approach, we also calculated the upper bound that the ensemble approach could achieve.  Specifically, as long as at least one model among the four pretrained language models correctly predicts the next word, we mark it as correct for the upper bound. This means that no other possible combination of the four pretrained language models can perform better. Here this upper bound is 66.44\% (Table \ref{tab:chapter5_results}), which is about a 2\% improvement over our ensemble approach; there is a bit of room for improvement, but also better language modeling techniques also need to be explored.  

Figure \ref{fig:ensemble_pos} shows the accuracy for RoBERTa, AutoMeTS, and the upper bound based on size of the context, i.e., words typed so far. For small context, the ensemble approach performs much better than RoBERTa.  This likely can be attributed to selecting one of the other models that performs better for small context, e.g., GPT-2.  As the context size increases, however, RoBERTa and the ensemble model perform similarly.  The upper bound is consistently above the ensemble approach across all context sizes.

\section{Conclusion \label{sec:chapter5_conclusion}}

In this chapter, we introduced a new medical parallel corpus, where a difficult sentence is paired with a corresponding simple sentence, for text simplification. This corpus contains 3.3K medical sentence pairs. Further, we proposed a novel autocomplete application of pretrained language models for medical text simplification. Such autocomplete models can assist users in simplifying text with improved efficiency and higher quality results in domains where information preservation is especially critical, such as healthcare and medicine, and where fully-automated approaches are not appropriate. We examined four recent pretrained neural language models: BERT, RoBERTa, XLNet, and GPT-2, and showed how data scarcity affected the model learning of NLP tasks. To offset the effect of data scarcity, we introduced AutoMeTS, an ensemble method that combines and leverages the partial learning of different individual pretrained neural language models. The AutoMeTS model outperforms the best individual model, RoBERTa, by 2.1\%. We also introduced a simple and model-agnostic solution to effectively incorporating the additional context of the sentences being simplified into the autocomplete simplification process.

\chapter{Conclusions and Future Work} 
\thispagestyle{fancy} 

In this dissertation, we presented our solutions to mitigating data scarcity using data augmentation and neural ensemble learning techniques for neural language models. In both research directions, we present simple and effective solutions and evaluate their impact on assisting neural language models in downstream NLP tasks. 

Specifically, for data augmentation, we explore two techniques: 1) context-based and 2) sequence-to-sequence using text simplification as a transformation method in low-resource NLP domains and tasks. In the former data augmentation technique, we generate additional training data by selectively moving a correct answer span and pairing newly created spans with the original question. Despite its simplicity, this method provides helpful information about the important context where the correct answer span is most likely to appear.  Our approach helps improve document retrieval performance, and when combined with the original answer span extraction component, outperforms the state-of-the-art on two real-world datasets in low-resource domains where creating additional training data is expensive. In the second data augmentation technique, we propose a novel use of text simplification to create additional training data for neural language models. Our approach introduces a variety of writing styles to training data and therefore, improves neural language models' ability to generalize. We showed that 1) with reasonable quality control, text simplification can be an effective solution to mitigate data scarcity in natural language processing and 2) neural language models still continue to significantly benefit from more training data, even in tasks with decent amounts of pre-existing training data. Across two data augmentation techniques, we showed that despite its effectiveness in understanding natural language, large pretrained models need a lot of annotated training data to learn a specific downstream task and data augmentation can be an off-the-shelf solution to quench this thirst for neural language models.

For neural ensemble learning, we trained a multi-label neural classifier to select the best prediction outcome from a variety of individual pretrained neural language models for autocomplete medical text simplification, a low-resource NLP task. We showed that with little training data, neural language models can only learn part of a natural language task. Synthesizing their learning by neural ensemble learning can be an effective solution to this problem. Our approach significantly improves general text generation performance on medical text simplification without the need for annotating more training data, which is expensive and often difficult to achieve, especially in medical domains. To further contribute to good progress of the text simplification research community, we also introduced a new medical parallel corpus, which contains 3.3K medical sentence pairs, for text simplification.

\begin{figure}
    \centering
    \scalebox{0.58}{\includegraphics{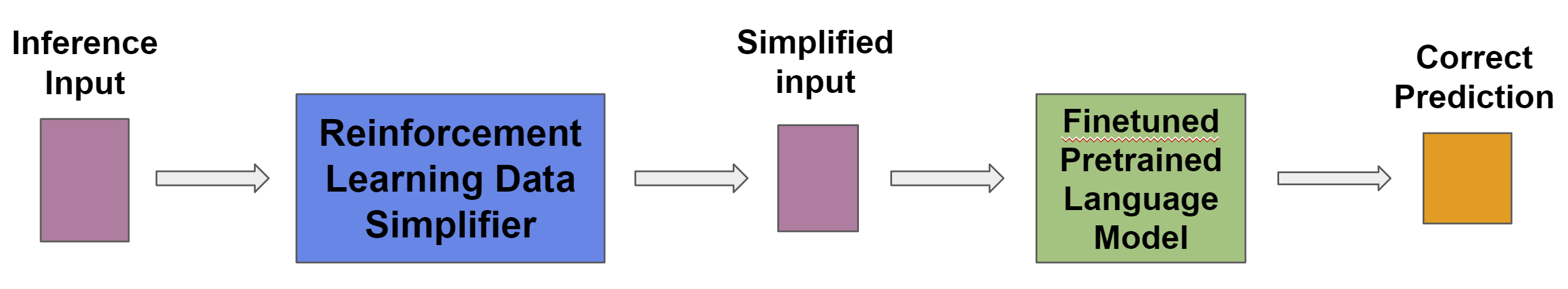}}
    \caption[Visualization of our proposed reinforcement learning method in simplifying text input for machine at inference time. Note that, we do not modify the underlying model trained for a downstream task.]{Visualization of our proposed reinforcement learning method in simplifying text input for machine at inference time. Note that, we do not modify the underlying model trained for a downstream task.}
    \label{fig:chapter6_reinforcement_learning}
\end{figure}

One future research direction that attracts our attention is to create a joint-learning method for simplifying text input for machines. In particular, such a method should be capable of directly incorporating model performance from downstream NLP tasks into the training process for text simplification. With recent advances in the field of natural language generation, we think that reinforcement learning is a strong candidate for this research direction. This is mostly because reinforcement learning encourages a strong machine-environment interaction and leverages it to regulate the machine learning process. Figure \ref{fig:chapter6_reinforcement_learning} shows our proposed use of reinforcement learning as a solution to this generation problem. Specifically, we propose to use the underlying neural language model that addresses a downstream task as an environment to provide direct rewards for a text simplification reinforcement learning agent. In this setting, a text simplification component can learn to simplify text specifically for an underlying model. That is, at each iteration, the text simplification reinforcement learning agent reads an inference input and takes a series of actions, e.g. words in a vocabulary, to generate the corresponding simplification. This simplification will then be input for the underlying model to make a prediction. The algorithm will reward the reinforcement learning agent if the prediction is correct and penalize it otherwise. An advantage of this approach is that the text simplification model can be trained end-to-end using direct instructions from the underlying model that is used for a downstream prediction task. Not only does this help to offset the need for the two-stage text simplification approach we introduced in chapter \ref{chapter:aug_data_ts}, but it also provides a way to incorporate other useful training objectives, e.g. control on sentence length, into the learning process.  

All in all, we are excited about making contributions to the field of data scarcity. With our findings in controlling text simplification quality, we hope to attract more attention in the AI community to create more fine-grained automated evaluation methods for text simplification for machine. Longer term, we envision that our unique take on autocomplete medical text simplification could encourage more researchers to work on other interesting applications and advancing text simplification in medical domains.



\chapter*{References}
\thispagestyle{fancy} 
\addcontentsline{toc}{chapter}{REFERENCES} 
\bibliography{bibliography} 

\end{document}